\documentclass[10pt,twocolumn,letterpaper]{article}
\usepackage[pagenumbers]{cvpr} 

\usepackage{amsmath,amsfonts,bm}

\def\eqref#1{equation~\ref{#1}}

\def\1{\bm{1}}

\DeclareMathAlphabet{\mathsfit}{\encodingdefault}{\sfdefault}{m}{sl}
\SetMathAlphabet{\mathsfit}{bold}{\encodingdefault}{\sfdefault}{bx}{n}

\def\gD{{\mathcal{D}}}

\def\gH{{\mathcal{H}}}

\def\gL{{\mathcal{L}}}

\def\gX{{\mathcal{X}}}
\def\gY{{\mathcal{Y}}}

\newcommand{\E}{\mathbb{E}}

\DeclareMathOperator*{\argmin}{arg\,min}

\newcommand{\braces}[2][]{#1\{#2 #1\}}

\usepackage{url}
\usepackage[T1]{fontenc}    
\usepackage{url}            
\usepackage{booktabs}       
\usepackage{amsfonts}       
\usepackage{nicefrac}       
\usepackage{microtype}      
\usepackage{multirow}
\usepackage{tikz}
\usetikzlibrary{calc}
\usepackage{comment}
\usepackage{caption}
\usepackage{subcaption}
\usepackage{graphicx}
\usepackage{soul}
\usepackage{amsmath, amsthm,amssymb}
\usepackage{pifont}
\usepackage{inconsolata}
\usepackage{tabto}
\usepackage[linesnumbered,ruled,vlined]{algorithm2e}
\usepackage {algpseudocode}
\usepackage{algorithmicx}
\usepackage{algcompatible}
\usepackage{xfrac}   
\usepackage{adjustbox}
\usepackage{wrapfig}
\usepackage[normalem]{ulem}
\usepackage{pgfplots}
\pgfplotsset{compat=1.7}
\usepackage{enumitem}
\usepackage[most]{tcolorbox}

\newcommand{\paragrapht}[1]{\noindent\textbf{#1}}

\usepackage{lipsum}

\definecolor{cvprblue}{rgb}{0.21,0.49,0.74}
\usepackage[pagebackref,breaklinks,colorlinks,allcolors=cvprblue]{hyperref}

\def\paperID{12181} 
\def\confName{CVPR}
\def\confYear{2025}

\title{PEER pressure: Model-to-Model Regularization for Single Source Domain Generalization}

\author{Dong Kyu Cho\\
New York University\\
New York, USA\\
{\tt\small dongkyu.cho@nyu.edu}
\and
Inwoo Hwang\thanks{Corresponding authors.}$\:\:$\thanks{Work done at Seoul National University.}\\
Columbia University\\
New York, USA\\
{\tt\small ih2455@columbia.edu}
\and
Sanghack Lee\footnote[1]{}\\
Seoul National University\\
Seoul, South Korea\\
{\tt\small sanghack@snu.ac.kr}
}

\begin{document}
\maketitle

\begin{abstract}

Data augmentation is a popular tool for single source domain generalization, which expands the source domain by generating simulated ones, improving generalization on unseen target domains. In this work, we show that the performance of such augmentation-based methods in the target domains universally fluctuates during training, posing challenges in model selection under realistic scenarios. We argue that the fluctuation stems from the inability of the model to accumulate the knowledge learned from diverse augmentations, exacerbating feature distortion during training. Based on this observation, we propose a novel generalization method, coined Parameter-Space Ensemble with Entropy Regularization (PEER), that uses a proxy model to learn the augmented data on behalf of the main model. The main model is updated by averaging its parameters with the proxy model, progressively accumulating knowledge over the training steps. Maximizing the mutual information between the output representations of the two models guides the learning process of the proxy model, mitigating feature distortion during training. Experimental results demonstrate the effectiveness of PEER in reducing the OOD performance fluctuation and enhancing generalization across various datasets, including PACS, Digits, Office-Home, and VLCS. Notably, our method with simple random augmentation achieves state-of-the-art performance, surpassing prior approaches on sDG that utilize complex data augmentation strategies.

\end{abstract}

\section{Introduction}
\label{sec:intro}
\begin{figure}
\centering
\includegraphics[width=.45\textwidth]{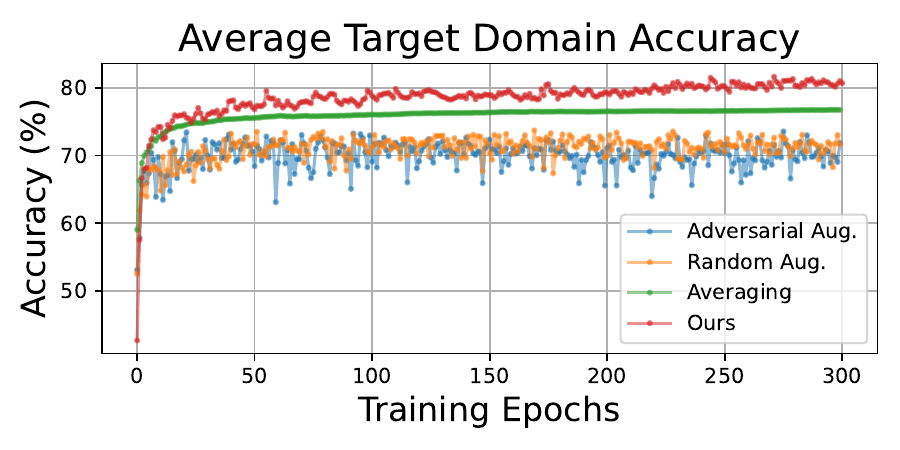} 
\caption{
Despite its generalization effect, data augmentation induces
fluctuations in target domain accuracy during the training.
This phenomenon becomes more pronounced as the complexity of the augmentation increases, complicating model selection. We address this issue of fluctuation with a simple model-to-model regularization method that cushions the effect of data augmentation.
}
\label{fig:what_is_fluctuation}
\end{figure}

Real-world deployment of deep neural networks frequently encounters {domain shift}, which refers to the discrepancy between the training domain and the unseen target domain on which the model is tested. An important aspect of domain shift is that it hinders the generalization of trained models \citep{kurakin2016}. Nevertheless, a trained model is expected to perform well on {various} OOD data, given a {limited} source of training data. Similarly, single source domain generalization (sDG) is the task of building a robust model that performs well across {multiple} target domains, trained from a {single} source domain \citep{dgsurvey2021}. Existing approaches commonly utilize data augmentation to generate simulated target domains \citep{volpi2018} and attempt to learn domain-invariant features from the augmented data.

\begin{figure}
    \centering
    \includegraphics[width=0.75\linewidth]{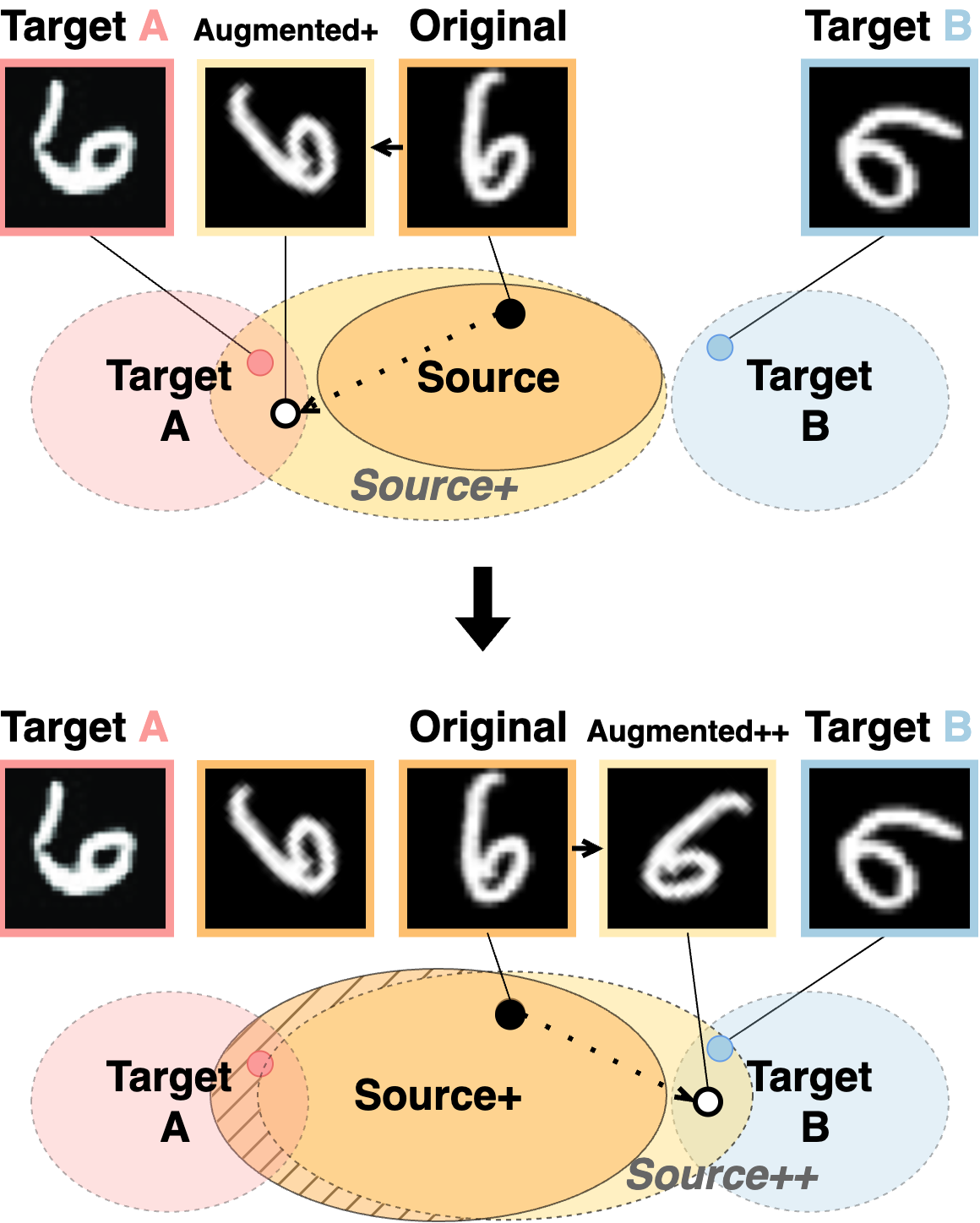}
    \caption{Illustration of pitfalls of augmentation in generalizing to unseen target domains. (a) Augmentation-based methods expand the source domain by providing diverse augmented samples (i.e., Source+). This enhances the model's generalization capability towards the unseen target domain (i.e., Target A). (b) Throughout the course of training, it iteratively simulates diverse unseen domains. However, at the same time, diverse augmentations lead to the distortion of the learned representations, thereby triggering OOD fluctuation.}\label{fig:pitfalls}
\end{figure}

This paper highlights an overlooked issue of leveraging data augmentation for sDG, particularly focusing on the fluctuation of OOD target domain performance amidst training, referred to as \textit{mid-train OOD fluctuation} (\cref{fig:what_is_fluctuation}). We find that this phenomenon stems from the model's incapability to accumulate the knowledge obtained from diverse augmentations and demonstrate that the features obtained from previous steps are largely distorted during training (see \cref{fig:pitfalls}). We further illustrate that the fluctuation worsens when the model's trained features are distorted by augmented samples discrepant from the previously trained data and show that augmented samples are surprisingly inconsistent from their original state. 
This complicates \textit{model selection} and potentially undermines generalization at test time, and thus, it is crucial to mitigate this issue.

Based on our observations, 
we suggest a novel generalization method coined \textsc{peer} (Parameter-Space Ensemble with Entropy Regularization), that 
mitigates the augmentation-induced feature distortion by averaging parameters at various points along the model's learning trajectory \citep{izmailov2018averaging}.
Specifically, our method leverages two interacting modules, i.e., the task model and the proxy model, to accumulate the knowledge acquired during training. 
The parameter-averaged task model guides the learning process of the proxy model, significantly reducing the aforementioned mid-train OOD fluctuation. Consequently, our framework stacks the generalization effect of varying data augmentation into the task model, reaching state-of-the-art performance in conventional sDG benchmarks (e.g., PACS, Digits), even in benchmarks where conventional sDG methods do not guarantee generalization (e.g., Office-Home, VLCS).

Our contributions are summarized as follows:
\begin{itemize}[leftmargin=*] 
\item We highlight an overlooked issue of the mid-train OOD fluctuation of augmentation-based sDG methods which poses serious issues in model selection and reveal that it stems from the distortion in the trained features. 
\item Based on our observation, we introduce \textsc{peer}, a novel framework for sDG that stabilizes the learning process and boosts the target domain accuracy by accumulating the generalization effect of diverse augmentations using a parameter-space ensemble model.
\item Our method achieves state-of-the-art performance across a wide range of benchmarks against existing sDG methods.
\end{itemize}

\section{Related Works}
\label{sec:preliminaries}

\paragraph{Domain generalization.}
In the multi-source domain generalization (DG) literature, learning domain-invariant features has shown success in training robust models \citep{IRM}. Specifically, these algorithms aim to disentangle the knowledge shared across domains \citep{klindt2020,ren2021}. A recent line of work highlighted the use of pre-trained models for model-to-model regularization, e.g., \citet{MIRO} used an external pre-trained model to encourage the learning of domain-invariant features, and \citet{li2023simple} expanded this approach by using multiple pre-trained models. In contrast, we refrain from using an external model and show that a training model can effectively perform regularization. On a different note, \citet{arpit2022ensemble} studied the instability of the model's OOD performance and suggested an ensemble algorithm to alleviate the stochastic nature of the learning process. In contrast, we relieve the computational burden of ensembles by using a single parameter-averaged model \citep{ainsworth2023git, rame2022diwa,jolicoeur2023population} and incorporate an alignment strategy \citep{choshen2022fusing,frankle2020linear} to assist this.

\paragraph{Single source domain generalization.}
In the sDG setting, only one domain is available for training, which makes it hard to apply conventional approaches developed for DG. To tackle this, a line of work focused on generating diverse
domains using sophisticated data augmentation strategies,
e.g., adversarial augmentation \citep{volpi2018} or
learnable augmentation modules \citep{fan2021,qiao2020,li2021,wang2021,xu2023simde,zheng2024advst}. 
On the other hand, we reveal a universal phenomenon (i.e., mid-train OOD fluctuation) associated with utilizing data augmentation for generalization, and present a simple strategy to alleviate it.

\paragraph{Mode connectivity and parameter-space ensembles.}
Our work draws inspiration from the mode connectivity \citep{frankle2020linear} property of neural networks, which refers to the presence of a continuous manifold of non-increasing error that connects the minima identified by two global minimizers (i.e., trained models) \citep{garipov2018loss,lubana2023mechanistic}. The concept is commonly used to justify how individual models can be merged to produce parameter-space ensembles \citep{wortsman2022model,rame2022diwa} and also form the basis for designing model alignment methods to encourage mode connectivity between models \citep{entezari2021role,choshen2022fusing, ainsworth2023git, rame2023model}. To analyze mode connectivity between models, a common practice is to measure the loss barrier \citep{frankle2020linear}, quantified as the rise in loss values when the parameters of two models are averaged. Extending this, we suggest an effective alignment method to encourage mode connectivity between models trained with varying augmented data.

\begin{figure*}[t!]
\centering
\begin{minipage}[t]{0.44\textwidth}
\centering
\captionof{table}{Empirical study of (a) target domain accuracy, (b) mid-train OOD fluctuation, and (c) source-target dataset distance. We use MNIST as a source. Large source-target distance (red) coincided with low target accuracy and high OOD fluctuation during training, and vice versa (blue).
} 
\fontsize{9}{10}\selectfont
\centering
\label{tab:gains_losses_da}
\begin{adjustbox}{width=0.8\textwidth}
\tabcolsep=0.11cm
\begin{tabular}{@{}c@{\hspace{0.5em}}lccccc@{}} 
    \toprule
    & {Method} & {SVHN} & {M-M} & {S-D} & {USPS} & Avg.\\
    \midrule
    &\multicolumn{6}{l}{(a) Target domain accuracy}\\
    \midrule
    & NoAug & \textcolor{BrickRed}{27.83} & 52.72 & 39.65& \textcolor{Cerulean}{76.94} & 49.29\\
    & RandAug [\citenum{cubuk2020randaugment}] & \textcolor{BrickRed}{57.76} &  77.15 & 73.65 & \textcolor{Cerulean}{87.94} & 73.98\\
    & AdvAug [\citenum{li2021}] & \textcolor{BrickRed}{62.21} & 82.20 & 69.39 & \textcolor{Cerulean}{85.26} & 74.77\\
    \midrule
    &\multicolumn{6}{l}{(b) Variance of the target domain accuracy}\\
    \midrule
    & NoAug & \textcolor{BrickRed}{4.76} & 2.77 & 1.72 & \textcolor{Cerulean}{0.32} & 1.33\\
    & RandAug [\citenum{cubuk2020randaugment}] & \textcolor{BrickRed}{2.51} & 1.04 & 1.05& \textcolor{Cerulean}{1.49} & 1.52 \\
    & AdvAug [\citenum{li2021}] & \textcolor{BrickRed}{3.58} & 2.56& 2.36 & \textcolor{Cerulean}{3.48} & 2.99\\
    \midrule
    &\multicolumn{6}{l}{(c) Source-target dataset distance [\citenum{alvarez2020geometric}] ($\times 10^3$)}\\
    \midrule
    \multirow{1}{*}{\rotatebox[origin=c]{90}{}}
    & -
    & \textcolor{BrickRed}{\textit{3.46}} & \textit{2.65} & \textit{2.75} & \textcolor{Cerulean}{\textit{0.92}} & \textit{2.45} \\ 
    \bottomrule
\end{tabular}    

\end{adjustbox}
\end{minipage}
\hfil
\begin{minipage}[t]{0.47\textwidth}
\centering
\vspace{0pt}
\includegraphics[width=0.6\textwidth]{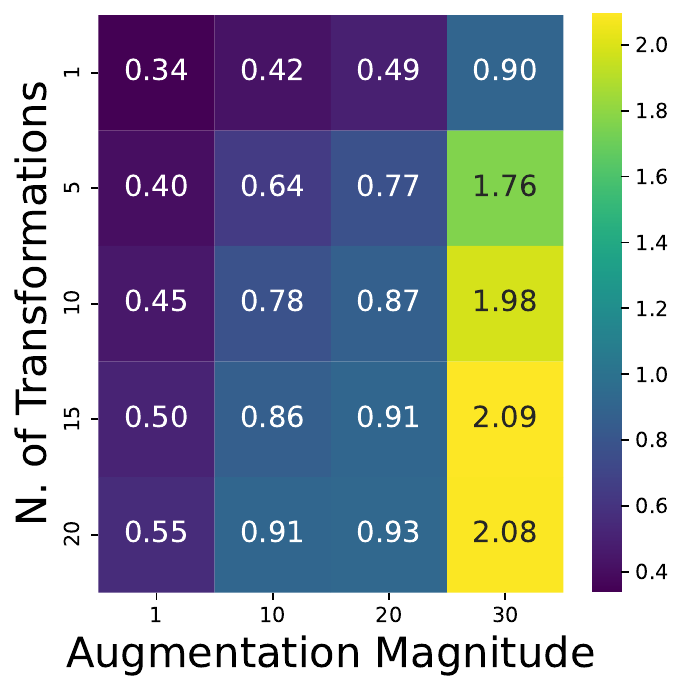}
\caption{OTDD distance [\citenum{alvarez2020geometric}] between the original data (MNIST) and its augmented view.} \label{fig:otdd_small}

\end{minipage}
\end{figure*}

\section{Observation: Pitfalls of Augmentation for Generalization}
\label{sec:limitation}

In this section, we reveal an overlooked problem in augmentation-based sDG methods. 
We first provide a brief background on the augmentation-based approaches to sDG (\cref{sec:3-1}).
Then, we highlight the performance fluctuation of models trained with data augmentation (\cref{sec:3-2}).

\subsection{Augment-and-Align: Augmentation-based Approaches to sDG}\label{sec:3-1}

Let $\gD_S=\{(x_i, y_i)\}^N_{i=1}$ be a source domain where $x_i\in\gX$ is an input image and $y_i\in\gY$ is its corresponding label.
The goal of sDG is to build a model $F$ from $\gD_S$
that is capable of generalizing to unknown target domains $\{\gD^{(1)}_T, \cdots, \gD^{(t)}_T\}$ distributionally different from the source domain.
The model $F=C\circ H$ consists of a feature extractor $H: \gX \to \gH$ and the classifier $C: \gH \to \gY$.
Clearly, the classifier relying on the domain-specific features would not generalize to unseen target domains, and thus it is crucial to learn domain-invariant features from the source domain. 

Existing approaches utilize data augmentation to simulate domain shift and aim to extract domain-invariant features by aligning the feature distribution between the original sample $x$ and its augmented view $\bar{x}=G(x)$, where $G$ is the augmentation function.
The objective of such augmentation-based sDG
approaches, omitting some arguments for simplicity, can be written as:
\begin{equation}
\argmin_{H, C} \E_{(x, y)\in \gD_S}  \Bigl( \gL_{\text{CE}}(C(H(x)), y) + \gL_{\text{align}}(x,\bar{x};H) \Bigr),
\label{eq:asDG} 
\end{equation}
where $\gL_{\text{CE}}$ is the cross-entropy loss
and $\gL_{\text{align}}$ is an alignment loss 
for capturing domain-invariant features by comparing $H(x)$ and $H(\bar{x})$. 
The commonly used alignment loss 
is InfoNCE \citep{oord2018}, which lower bounds the mutual information $I(H(x), H(\bar{x}))$.
Importantly, such alignment only guarantees to retrieve \textit{augmentation-invariant} features \citep{ssl_separates}, and simple input transformations for generating the augmented views are often insufficient to capture \textit{domain-invariant} ones \citep{aminbeidokhti2023domain}.
Therefore, recent methods devise more complex data augmentation strategies \citep{wang2021,li2021} to simulate diverse shifts in distribution.

However, it is still unclear whether such augmentation strategies can guarantee generalization to the target domain, especially given that it is unseen. In the sequel, we illustrate that this discrepancy makes the model performance fluctuate in the target domain.

\subsection{Mid-train OOD fluctuation of Augmentation-Based sDG Methods}\label{sec:3-2}

Recall \cref{fig:what_is_fluctuation}, we find that augmentation-based sDG methods commonly exhibit large fluctuation of OOD performance throughout training, dubbed {mid-train OOD fluctuation}. 
Then, the following questions naturally arise: \textit{``How does the fluctuation relate to the generalization performance? Where does the fluctuation stem from?''}
Here, we investigate the relationships between the fluctuation and target domain accuracy through the lens of source-target dataset distance and examine the impact of data augmentation on the fluctuation.

We begin by observing that the target domain accuracy is closely related to the mid-train OOD fluctuation by comparing two augmentation-based sDG methods: random augmentation (RandAug, \citet{cubuk2020randaugment}) and adversarial augmentation (AdvAug, \citet{li2021}).
As shown in the last column (Avg.) of \cref{tab:gains_losses_da}-(a) and (b), the models with better generalization performance also display larger fluctuation. 
Clearly, the complexity of the augmentation the models employed aligns with the target domain accuracy and fluctuation.

To further investigate their relationships, we adopt a similarity metric that measures the geometric distance between datasets  (i.e., OTDD \citep{alvarez2020geometric}).
By comparing different target domains (e.g., SVHN and USPS), we observe that the source-target discrepancy shown in \cref{tab:gains_losses_da}-(c) is closely associated with the target domain accuracy and fluctuation.
In other words, the models exhibit relatively small fluctuation on the target domain that is similar to the source domain (i.e., USPS) and vice versa (i.e., SVHN).
Similarly, the models tend to show higher accuracy on target domains with smaller discrepancies (i.e., USPS) and vice versa (i.e., SVHN).

To better understand our observations above, we examine the discrepancy between the original dataset (MNIST) and its augmented view across varying degrees of random augmentation \citep{cubuk2020randaugment}.
As shown in \cref{fig:otdd_small}, we observe that the discrepancy becomes more significant as the augmentation becomes diverse and its magnitude becomes stronger.
Notably, such discrepancies often even exceed the source-target distance (i.e., $0.92$ in \cref{tab:gains_losses_da}-(c)). 

Lastly, we find that parameter-averaging of multiple points along the model's learning trajectory \citep{izmailov2018averaging} can drastically reduce the OOD fluctuation, although with only limited gains in generalization. This is illustrated by the green line in \cref{fig:what_is_fluctuation}. Intuitively, this aligns with our idea that the model's learned features are consistently distorted during training, and parameter-averaging could alleviate the distortion \citep{marouf2023weighted}. 

Our observations suggest that data augmentation improves generalization capacity by simulating diverse domain shifts, but simultaneously leads to the distortion of the learned features and triggers mid-train OOD fluctuation, as depicted in \cref{fig:pitfalls}. 
Based on our findings, we now proceed to present our method that retains the knowledge accumulated throughout the training, thereby alleviating fluctuations while achieving better generalization performance.

\section{Method}
\label{sec:method}
We now present a novel generalization method for sDG, coined Parameter-Space Ensemble with Entropy Regularization (\textsc{peer}), that mitigates the augmentation-induced feature distortion and its associated issues (e.g., mid-train OOD fluctuation).
Our approach involves two interacting modules with identical architectures: a frozen task model $F$ and a trainable proxy model $P$. The task model guides the proxy model's learning process through entropy regularization of feature representations (\cref{sec:peer_regularization}). Subsequently, the task model is updated via parameter-averaging with the regularized proxy model, progressively accumulating the proxy model's knowledge throughout training (\cref{sec:peer_update}). The concept of our method is depicted in \cref{fig:framework}. The pseudo-code of our method is provided in \cref{alg:peer}.

\subsection{Regulating the Proxy Model with \textsc{peer}}\label{sec:peer_regularization}

Our goal is to learn a robust task model $F$ from a single source domain that can generalize to multiple unseen target domains, where the task model consists of a frozen encoder $H_{f}: \mathcal{X} \rightarrow \mathcal{H}$ and a frozen classification head $C_f: \mathcal{H} \rightarrow \mathcal{Y}$, i.e., $F={C_f}\circ{H_f}$. However, directly training the task model with varying augmented data is prone to feature distortion. Our key idea is to introduce a proxy model $P$ that trains on behalf of the task model and under the its guidance.
Specifically, the proxy model $P= C_p\circ H_p$ shares the same architecture as the task model and consists of an encoder $H_p: \mathcal{X} \rightarrow \mathcal{H}$ and a classification head $C_p: \mathcal{H} \rightarrow \mathcal{Y}$. The proxy model is initialized by copying the task model at the beginning of training, i.e., $\theta_p \gets \theta_f^{(0)}$ where $\theta_p$ is the parameters of the proxy model $P$ and $\theta_f^{(n)}$ is the parameters of the task model $F$ at $n$-th training epoch.

Our method \textsc{peer} imposes regularization to the proxy model at the intermediate feature level. Instead of directly comparing the intermediate representation in $\mathcal{H}$, we map the representations from $H_f$ and $H_p$ using a shared projection head $R: \mathcal{H} \rightarrow \mathcal{R}$, 
following the empirical analysis by \citet{gupta2022understanding} and our experimental findings (\cref{tab:accuracy_projection_head}) regarding its optimization efficacy.

The objective for \textsc{peer} is then defined as: 
\begin{align} 
\gL_{\textsc{peer}}(H_{f}(x), H_{p}(\bar{x})) = - I(R(H_{f}(x)); R(H_{p}(\bar{x}))), \label{loss:peer} 
\end{align} 

where $x$ denotes the original sample and $\bar{x}$ the augmented view created by an augmentation function $G$. The loss function $\gL_{\textsc{peer}}$ is designed to maximize the mutual information ($I$) between the two representations $R(H_{f}(x))$ and $R(H_{p}(\Bar{x}))$. Since the exact mutual information is intractable, we use practical lower bounds $\Tilde{I}$ such as the InfoNCE \citep{oord2018} or the Barlow Twins \citep{barlow_twins} loss functions, both effective in optimizing mutual information between feature representations \citep{poole2019variational}. The details of the mutual information optimization are included in \cref{appendix:mi_optimization}.

\begin{algorithm}[t]
\caption{Parameter-space Ensemble with Entropy Regularization (\textsc{peer})}
\label{alg:peer}
\footnotesize
\textbf{Input:} Task model $F$ and its parameter $\theta_f$, augmentation function $G$, data from source domain $D_{s}$,
augmentation reinitialization criteria $k$;  

\textbf{Output:} Fully updated task model $F$ and its parameter $\theta_f$

Pre-train $F$ with $D_s$ without $G$

Initialize $P$ by setting its parameter $\theta_{p}$ with $\theta_f$ from $F$

Initialize trajectory $\Theta \gets \{\ \}$ 

\While{not converge}{
    \If{ $n\mathbin{\%}k$=$0$ }{
    Reinitialize $G$ \tabto{3cm}\tcp{for random augmentation, change augmentation strength}
    $\Theta \gets \Theta \cup \{\theta_p^{(n)}\}$ \tabto{3cm}\tcp{save a snapshot of $P$}
    $\theta_{f} \gets \textsc{average}(\Theta)$ \tabto{3cm} \tcp{update $F$ \textrm{(\Cref{eq:peer_update})}}
    }
    \For{$i=1 : n_{iterations}$}{
     

     Augment the $i$-th mini-batch sampled from $D_s$ with $G$
     
     Train $P$ with \textsc{peer} following \Cref{loss:f} 
    }
}

\end{algorithm}


Intuitively, regularizing with \textsc{peer} guides the proxy model $P$ to learn features selected by the task model $F$. Notably, in \cref{loss:peer}, the task model and the proxy model receive nonidentical inputs $x$ and $\bar{x}$, respectively, reflecting our idea that the frozen task model is expected to provide a rich feature representation of the original sample $x$, while the training proxy model can better comprehend the newly augmented sample $\bar{x}$.

We train \textit{only} the proxy model $P$ using a classification loss (i.e., cross-entropy) with the regularization:
\begin{align}\label{loss:f}
\gL_{P} = \sum\nolimits_{x'\in \{x,\Bar{x}\}} &\gL_{\text{CE}}(C_p(H_p(x')),y) \\ \notag
+  &w \cdot 
\gL_{\textsc{peer}}(H_{f}(x),H_{p}(\Bar{x})), 
\end{align}
where $w$ is a balancing coefficient. In \cref{sec:discussion_peer}, we further elaborate on the \textsc{peer} regularization as an optimization of the mutual information (MI).

\subsection{Accumulating Knowledge in the Task Model with \textsc{peer}}\label{sec:peer_update}

The task model $F$ is gradually updated through parameter-averaging with the proxy model $P$.
This updating process progressively improves the task model's generalization throughout training, ensuring it remains effective as the regulator of the ever-growing proxy model \citep{burns2023weaktostrong}. 
Specifically, we update the task model by parameter-averaging with the proxy model for every $k$ epoch through the proxy model's learning trajectory i.e., $\Theta= \braces[\big]{\theta_p^{(k)}, \theta_p^{(2k)} \cdots, \theta_p^{(\lfloor \frac{n}{k} \rfloor \cdot k)}}$ where $n$ is the current training epoch, and update the task model with:
\begin{equation}\label{eq:peer_update}
\theta_f \gets \frac{1}{\lvert \Theta \rvert}\sum_{\theta \in \Theta} \theta.
\end{equation}
Also, we reinitialize the augmentation function $G$ for every $k$ epoch (e.g., changing the policy -- number and of transformations/ magnitude --  of random augmentation). This periodic update of the task model allows it to stack the effect of diverse augmentations, similar to an ensemble model \citep{rame2022diwa}.

For the parameter-averaged task model to enjoy ensemble effects, it's crucial to ensure mode connectivity \citep{frankle2020linear} between the task model and the proxy model, which can be sufficed by sharing an identical initialization or backbone \citep{neyshabur2020being}. As our proxy model is initialized from the task model, it naturally satisfies this requirement. To further benefit parameter-averaging, the two models must be closely located in the feature space, which can be obtained by tuning the models on an identical source data \citep{rame2023model,choshen2022fusing}. Our regularization with \textsc{peer} (\cref{loss:peer}) encourages the proxy model to be aligned with the task model in the feature space by treating the augmented domain similarly to the source domain. In \cref{sec:experiment}, we show that the task model and the proxy model benefit from the regularization's alignment effect. In \cref{appendix:model-to-model}, we empirically demonstrate that the task model cannot function as an effective regulator of the proxy model without the updating process (w/o ParamAvg. in \cref{tab:peer_vs_teacher}).

\subsection{Discussion}\label{sec:discussion_peer}

\paragrapht{\textsc{peer} as mutual information (MI) maximization.}
The idea of \textsc{peer} is that we can leverage the frozen task model to regularize the proxy model by maximizing the shared information between the two models. \textsc{peer} aims to maximize the MI between the intermediate output features of the two encoders $H_f$ and $H_p$. The entropy regularization aligns the proxy model to the task model, preventing the proxy model from deviating too far from the task model. From this perspective, an intended objective for \textsc{peer} could be formulated as 
$\max_{H} \  I(H_{f}(\bar{x});H_{p}(x))$
where $I(X;Y)= \mathbb{E}_{p(x,y)}[ \log {p(x \mid y)}/{p(x)}]$ indicates the mutual information i.e., MI. In our implementation, \textsc{peer} uses a feature decorrelation loss \cref{loss:bt} \citep{barlow_twins} to maximize the lower bound of MI as a surrogate objective for MI optimization under a Gaussian assumption \citep{tsai2021}. We further elaborate on the adequacy of \cref{loss:bt} for MI optimization in \cref{appendix:model-to-model} and report comparative results of different objectives e.g., InfoNCE \citep{oord2018} and Barlow Twins \citep{barlow_twins} (\cref{tab:table_infonce_bt}). In \cref{sec:experiment}, we provide experimental analysis on the effect of \textsc{peer} by showing its effectiveness in alleviating augmentation-induced feature distortion.

\section{Experiment}
\label{sec:experiment}

In this section, we investigate the following questions:
(1) How effective is our method compared to prior sDG approaches? (\cref{tab:all_sdg_conventional,tab:all_sdg_new})
(2) Does our method reduce the fluctuation of OOD performance? (\cref{tab:fluctuation_all})
(3) What effect does our method have on the model's learned features and loss landscape connectivity? (\cref{fig:lmc_alignment,fig:cka_full,fig:cka_peer_task})
(4) How effective is our method compared to previous model-to-model regularization approaches (\cref{tab:peer_vs_teacher}) or ensemble methods (\cref{tab:ensemble_w_wo})?

\subsection{Experimental Setup}\label{sec:setup}

\paragraph{Datasets.}
Following prior works \citep{li2021,wan2022}, we evaluate our method on two standard benchmarks for sDG.
\textbf{{PACS}} \citep{pacs} consists of 4 domains of differing styles (Photo, Art, Cartoon, and Sketch) with 7 classes. By default, we train our model with the Photo domain and evaluate it on the remaining target domains.
\textbf{Digits} comprises of 5 different digit classification datasets, MNIST \citep{deng2012mnist}, SVHN \citep{svhn}, MNIST-M (M-M) \citep{mnistm}, SYNDIGIT (S-D) \citep{syndigit}, and USPS \citep{usps}. We train our model with the first 10,000 samples of the MNIST dataset and assess its generalization accuracy across the remaining domains. 

We also include Office-Home \citep{officehome} and VLCS \citep{fang2013unbiased}, challenging benchmarks for sDG methods.
\textbf{Office-Home} is a common multi-DG benchmark consisting of 4 datasets (Real-world, Art, Clipart, Product) with differing styles with 65 classes. We train on the Real-world domain and evaluate with the remaining domains. 
\textbf{VLCS} is also a benchmark for multi-DG, comprised of 4 datasets, PASCAL-VOC (V), LabelMe (L), Caltech-101 (C), and SUN09 (S) with varying styles. We used the PASCAL-VOC dataset as the source and the rest as target domains. 

\vspace{-3mm}

\paragraph{Baselines.}
We first consider ERM \citep{ermkolt} and also compare our method with several strong augmentation-based approaches, i.e., M-ADA \citep{qiao2020}, L2D \citep{wang2021}, PDEN \citep{li2021}, SimDE \citep{xu2023simde} and AdvST \citep{zheng2024advst}. Some recent works \citep{xu2023simde,zheng2024advst} have reported the results using a different backbone (ResNet-18 in PACS) from the standard setting (AlexNet), thus we have used the authors' codes (if applicable) for reassessment. 

\vspace{-3mm}

\paragraph{Implementation.}\label{sec:implementation}
We use the same backbone architecture as prior works to ensure fair comparison. Specifically, we used AlexNet and multi-layer CNN for PACS and Digits, respectively, following earlier works \citep{wan2022,li2021,volpi}. For Office-Home and VLCS, we used ResNet-18. Additional experimental results across various backbone models (e.g., ResNet-18/50) are provided in Appendix (\cref{tab:accuracy_modelsize,tab:fluctuation_modelsize}). 
For the implementation of our method, we use random augmentation \citep{cubuk2020randaugment} to generate augmented samples. 
We set $k=10$ and the balancing coefficients $\lambda = 0.005$, and $w=2$ for all experiments.
Hyperparameter studies are provided in \cref{appendix:hyperparameter_experiment}.
We report the final test accuracy of the task model and report the OOD fluctuation measured as the variance of the target domain accuracy for every $k$-th epoch (\cref{tab:fluctuation_all}).
Throughout this section, we use the abbreviation RA for Random Augmentation and P for \textsc{peer}.

\subsection{Main Results}

\begin{table}[t!]
\caption{Target domain accuracy on PACS and Digits ($^\dagger$ indicates numbers are from original authors).}
\label{tab:all_sdg_conventional}
\centering
\begin{adjustbox}{width=0.48\textwidth}
\centering\begin{tabular}{@{}lccccccccc@{}}
\toprule
& \multicolumn{4}{c}{{PACS}} & \multicolumn{5}{c}{{Digits}}\\
\cmidrule(lr){2-5} \cmidrule(lr){6-10}
{Method} & A & C & S & Avg. & SVHN & M-M & S-D & USPS & Avg.\\
\midrule
ERM [\citenum{ermkolt}] & 54.43 & 42.74 & 42.02 & 46.39 & 27.83 & 52.72 & 39.65 & 76.94 & 49.29\\
ADA$^\dagger$ [\citenum{fan2021}] & 58.72& 45.58& 48.26& 50.85 & 35.51 & 60.41 & 45.32 & 77.26 & 54.62 \\
M-ADA$^\dagger$ [\citenum{qiao2020}] & 58.96 & 44.09 & 49.96 & 51.00 & 42.55 & 67.94 & 48.95 & 78.53 & 59.49\\
L2D$^\dagger$ [\citenum{wang2021}] & 56.26 & 51.04 & 58.42 & 55.24 & 62.86 & 87.30 & 63.72 & 83.97 & 74.46\\
PDEN [\citenum{li2021}] & 57.41 & 45.77 & 65.01 &  56.06 & 62.21 & 82.20 & 69.39 & 85.26 & 74.77\\
SimDE$^\dagger$ [\citenum{xu2023simde}] & -- & -- & -- & 59.32 & 66.08 & 84.90 & 70.04 & 86.56& 76.89 \\
AdvST [\citenum{zheng2024advst}] & 53.95 & 46.11 & 49.63 & 49.90 & 67.50 & 79.80 & 78.10 & 94.80 & 80.10 \\ 
MetaCNN$^\dagger$ [\citenum{wan2022}] & 54.05 & \textbf{53.58} & 63.88 & 57.17 & 66.50 & \textbf{88.27} & 70.66 & 89.64 & 78.76\\
\midrule
RandAug [\citenum{cubuk2020randaugment}] & 54.17 & 47.48 & 65.11 & 55.59 & 57.76 & 77.15 & 73.65 & 87.94 & 73.98\\
PEER (ours) & \textbf{62.66} & 47.40 & \textbf{68.21} & \textbf{59.42} & \textbf{70.79} & 76.84 &  \textbf{83.05} &  \textbf{93.57} &  \textbf{81.06}\\

\bottomrule
\end{tabular}

\end{adjustbox}
\end{table}

\begin{table}[t!]
\caption{Target domain accuracy on Office-Home and VLCS.}
\label{tab:all_sdg_new}
\centering
\begin{adjustbox}{width=0.48\textwidth} 
\centering\begin{tabular}{@{}lcccccccc@{}}
\toprule
& \multicolumn{4}{c}{{Office-Home}} & \multicolumn{4}{c}{{VLCS}}\\
\cmidrule(lr){2-5} \cmidrule(lr){6-9}
{Method} & Art & Clipart & Product & Avg. & L & C & S & Avg.\\
\midrule

ERM [\citenum{ermkolt}] & 52.78 & 40.19 & 68.73 & 53.90 & 59.06 & 97.30 & \textbf{74.25} & 76.87\\

M-ADA [\citenum{qiao2020}] & 54.36 & 40.41& 65.11 & 53.29 & 57.84 & 97.88 & 64.42 & 73.38\\
L2D [\citenum{wang2021}] & 54.02 & 41.77 & 66.30 & 54.03 & 56.21 & 95.52 & 66.90 & 72.87\\ 

PDEN [\citenum{li2021}] & 53.39 & 43.38 & 66.25 & 54.34 & 62.55 & 96.11 & 73.52 & 77.39\\
      
\midrule
      
RandAug [\citenum{cubuk2020randaugment}] & 43.10 & 45.47 & 61.67 & 50.01 & 57.58 & 93.18 & 66.56 & 72.44\\
      
PEER (ours) & \textbf{56.81} & \textbf{54.23} & \textbf{70.84} & \textbf{60.63} & \textbf{67.00} & \textbf{97.73} & 72.56 & \textbf{79.10}\\

\bottomrule
\end{tabular}

\end{adjustbox}
\end{table}

In \cref{tab:all_sdg_conventional,tab:all_sdg_new}, we report experimental results using the accuracy for each target domain and the mean accuracy across all target domains. In standard sDG benchmarks (i.e., PACS, Digits; \cref{tab:all_sdg_conventional}), our method achieves state-of-the-art target domain accuracy in many of the target domains and outperforms all baselines in terms of mean accuracy. Please note that SimDE and AdvST have used more robust backbones (ResNet-18) than the standard setting (AlexNet), which makes direct comparison challenging.
Notably, our method outperforms current SoTA methods (using the same backbone)
by $2.30\%$ and $0.96\%$. It is worth noting that our simple method boosted the mean accuracy of random augmentation (RandAug) by $7.08\%\uparrow$ in Digits and $3.76\%\uparrow$ in PACS. 

In more challenging benchmarks (i.e., Office-Home, VLCS; \cref{tab:all_sdg_new}),
previous augmentation-based methods (e.g., PDEN, RandAug) show either small gains or negative effects in enhancing generalization.
Similarly, naively applying random augmentation for these benchmarks lowered the target domain accuracy. In contrast, applying random augmentation with \textsc{peer}, shows a significant performance gain of $10.62\%$ in Office-Home and $6.66\%$ in VLCS.

\begin{table*}[t!]
\caption{Variance of the target domain accuracy.}
\label{tab:fluctuation_all}
\centering
\begin{adjustbox}{width=0.88\textwidth}\setlength{\tabcolsep}{4pt}
\centering

\begin{tabular}{@{}lccccccccccccccccc@{}}
\toprule
& \multicolumn{4}{c}{{PACS}} & \multicolumn{5}{c}{{Digits}} & \multicolumn{4}{c}{{Office-Home}} & \multicolumn{4}{c}{{VLCS}}\\
\cmidrule(lr){2-5} \cmidrule(lr){6-10} \cmidrule(lr){11-14} \cmidrule(lr){15-18}
{Method} & A & C & S & Avg. & SVHN & M-M & S-D & USPS & Avg. & Art & Clipart & Product & Avg. & L & C & S & Avg.\\
\midrule

L2D [\citenum{wang2021}] & 3.70 & 5.30 & 13.37 & 7.46 & 3.53 & 3.01 & 2.59& 4.44 & 3.39 & 5.22& 1.90& 5.58& 4.23& 5.72& 0.59& 1.66& 2.66\\

PDEN [\citenum{li2021}] & 3.39& 5.22& 7.23 & 5.28 & 3.58 & 2.56 & 2.36 & 3.48 & 2.99 & 10.63 & 2.17 & 7.46 & 6.75& 2.44 & 2.39& 2.81& 2.55\\

RandAug [\citenum{cubuk2020randaugment}] & 2.23 & 4.81 & 5.01 & 4.02 & 2.51 & \textbf{1.04} & 1.05 & 1.49 & 1.52 & \textbf{3.49} & 2.17 & 2.74 & 1.89 & 3.02 & 1.61 & \textbf{1.96} & 2.20\\

PEER (ours) & \textbf{2.01}  & \textbf{3.98} & \textbf{4.77} & \textbf{3.59} & \textbf{2.03} & 1.11 & \textbf{1.04} & \textbf{1.24} & \textbf{1.36} & 3.99 & \textbf{1.41} & \textbf{1.80} & \textbf{1.31} & \textbf{2.05} & \textbf{1.61} & 2.10 & \textbf{1.92}\\

\midrule
Metric & \multicolumn{16}{c}{Source-target dataset distance ($\times 10^3$)}\\ 
\midrule
OTDD [\citenum{alvarez2020geometric}] & 13.37 & 29.52 & 49.94 & 30.94 & 3.46 & 2.65 & 2.75 & 0.92 & 2.45 & 19.53 & 19.29 & 20.63 & 19.82 & 11.79 & 10.14 & 11.77 & 11.23 \\

\bottomrule
\end{tabular}

\end{adjustbox}
\end{table*}

Finally, \cref{tab:fluctuation_all} demonstrates the 
fluctuation of OOD performance, measured as the variance across the target domain accuracy. 
We observe that our method successfully reduces the mid-train OOD fluctuation across all benchmarks. 
In our framework, the task model accumulates knowledge of the proxy model throughout the training.
Thus, regularizing with the task model encourages the proxy model to preserve the knowledge of previous steps, similar to a memory buffer used in continual learning \citep{wang2024comprehensive}. In the next section, we illustrate that the task model indeed preserves the knowledge of the proxy model through parameter averaging.

\subsection{Detailed Analysis on \textsc{peer}}\label{sec:study_on_peer}

\subsubsection{Advantages of \textsc{peer} in Model-to-Model Regularization}\label{sec:exp_m2m}

In \cref{tab:peer_vs_teacher}, we demonstrate the advantages of \textsc{peer} compared to previous approaches that utilize a pre-trained model (i.e., teacher) for regularization,
where $\textsc{t+ra}$ and $\textsc{p+ra}$ refer to applying the teacher and the \textsc{peer} regularization, respectively. 
We observe that both the teacher and the task model in \textsc{peer} reduce the OOD fluctuation, while the fully-trained teacher ($\textsc{t+ra}$) often displays a stronger regularization effect compared to \textsc{peer} ($\textsc{p+ra}$). However, \textsc{peer} achieves superior sDG target domain accuracy in both datasets compared to the teacher.
This is due to the teacher model's static nature, which limits its capability to process newly augmented samples. In contrast, our task model, evolving with the proxy model, is less vulnerable to these limitations.  

We further validate the effectiveness of the updating process by 
ablating
parameter-averaging (w/o ParamAvg. in \cref{tab:peer_vs_teacher}). Instead of updating the task model by parameter-averaging, we simply freeze a snapshot of the proxy model for every $k$ epoch and use the latest snapshot as the regulator. 
As shown in \cref{tab:peer_vs_teacher}, the non-averaged task model sacrifices the target domain accuracy for addressing OOD fluctuation, which illustrates the effectiveness of parameter-averaging.

\begin{table*}[t!]
\caption{Comparitive study on \textsc{peer} vs. Teacher.}
\label{tab:peer_vs_teacher}
\centering
\begin{adjustbox}{width=0.8\textwidth}\setlength{\tabcolsep}{4pt}
\centering\tabcolsep=0.13cm
\begin{tabular}{@{}llccccccccc@{}} 
\toprule
& & \multicolumn{4}{c}{{PACS}} & \multicolumn{5}{c}{{Digits}}\\
\cmidrule(lr){3-6} \cmidrule(lr){7-11}
\textbf{Method} & Regulator & A & C & S & Avg. & SVHN & M-M & S-D & USPS & Avg.\\
\midrule
\multicolumn{11}{c}{Variance of the target domain accuracy (OOD Fluctuation)}\\
\midrule
RandAug [\citenum{cubuk2020randaugment}] & N/A & 2.23 & 4.81 & 5.01& 4.02 & 2.51 & 1.04 & 1.05 & 1.49 & 1.52\\
\textsc{t+ra} & Teacher  & \textbf{1.27} & \textbf{2.49} & 5.30 & \textbf{3.02} & 1.95 & 1.17 & \textbf{1.10} & \textbf{1.11} & \textbf{1.33}\\
\textsc{p+ra} & \textsc{peer} (w/o ParamAvg.)  & 1.69 & 3.38 & \textbf{4.62} & 3.23 & \textbf{1.93} & \textbf{1.10} & 1.11 & 1.22  & 1.34 \\ 
\textsc{p+ra} & \textsc{peer}  & 2.01 & 3.98 & 4.77& 3.59 & 2.03 & 1.11 & 1.04 & 1.24 & 1.36\\
\midrule
\multicolumn{11}{c}{Target Domain Accuracy}\\
\midrule
RandAug [\citenum{cubuk2020randaugment}] & N/A & 54.17 & \textbf{47.48} & 65.11& 55.59 & 57.76 & 77.15 & 73.65 & 87.94 & 73.98\\
\textsc{t+ra} & Teacher & 58.61 & 46.66 & 64.23 & 56.50 & 63.37 & 72.63 & 77.91 & 87.39 & 75.33 \\
\textsc{p+ra} & \textsc{peer} (w/o ParamAvg.)  & 57.73  & 46.69 & 61.33 & 55.25 & 59.99 & \textbf{77.26} & 72.3 & 88.28 & 74.46\\ 
\textsc{p+ra} & \textsc{peer} & \textbf{62.66} & 47.40 & \textbf{68.21} & \textbf{59.42} & \textbf{70.79} & 76.84 &  \textbf{83.05} &  \textbf{93.57} &  \textbf{81.06}\\
\bottomrule
\end{tabular}
\end{adjustbox}
\end{table*}

\subsubsection{Effect of \textsc{peer} on Parameter-Averaging}\label{sec:effect_paramavg}

\begin{figure}[tb]
\centering
\includegraphics[width=.4\textwidth, clip, trim= 0 1.05cm 0 0]{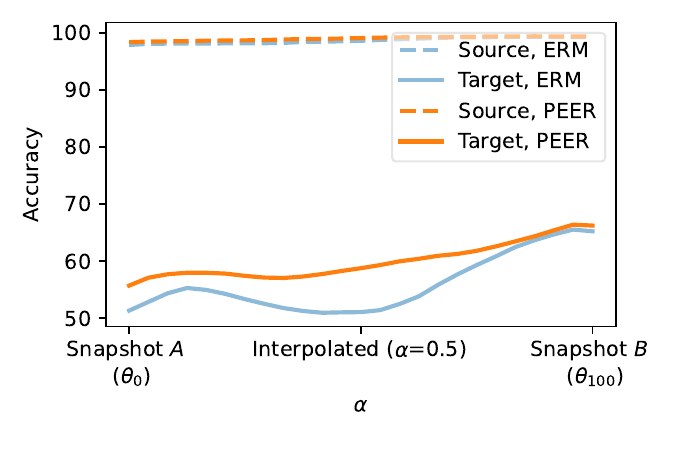}
\caption{Mode connectivity in the proxy model's trajectory. \textsc{peer} benefits parameter-averaging between snapshots of $P$ through its regularization effects.} \label{fig:lmc_alignment}
\end{figure}

Here, we investigate the effect of \textsc{peer} regularization in benefiting parameter-averaging for the task model $F$ update. We observe that the regularization brings forth an alignment between different steps of the proxy model in its learning trajectory $\Theta$. To clarify, we find different steps of the proxy model
$\theta_p^{(i)}, \theta_p^{(j)}$
to be aligned by the regularization. To show this, we follow the practice of \citet{frankle2020linear} and analyze the loss barrier between snapshots of the proxy model in its learning trajectory. \cref{fig:lmc_alignment} illustrates the mode connectivity of the proxy model training with data augmentation with/without \textsc{peer} on Digits (source: MNIST, target: SVHN). Here, we analyze the connectivity of the proxy model in its early stage of training ($\theta_p^{(0)}$) and at the late stage ($\theta_p^{(100)}$) by interpolating the two $\alpha\theta_p^{(0)} + (1-\alpha)\theta_p^{(100)}$, where $\alpha \in [0,1]$ be the interpolation weight. We note that \textsc{peer} aligns the model's snapshots($\theta_p^{(0)}, \theta_p^{(100)}$) in its learning trajectory, gifting a stronger performance gain when it is interpolated ($\alpha=0.5$), especially in the OOD target domain. In other words, \textsc{peer}'s regularization enables the task model to function as a robust parameter-space ensemble, which can guide the proxy model's generalization to target domains.

We further investigate the \textsc{peer}'s role in parameter-averaging in \cref{tab:ensemble_w_wo}, specifically showing the failure cases of parameter-averaging without model alignment. Here, P-ENS refers to the parameter-space ensembles. In both PACS and Digits, parameter-space ensembling without regularization (P-ENS w/o \textsc{peer}) falls behind ensembling with regularization. Notably in PACS, we observe failure cases of parameter-space ensembling without regularization, where the ensemble effect (i.e., gain in generalization ability) was very marginal. This failure case in parameter-averaging is an interesting observation as averaging the parameters between different training step snapshots of the same model has shown great success in many previous works \citep{grill2020bootstrap,izmailov2018averaging}. In \cref{appendix:mode_connectivity}, we provide a deeper analysis of this topic.

\begin{figure*}[tb]
     \centering
     \hfill
     \begin{subfigure}[b]{0.23\textwidth}
         \centering
         \includegraphics[clip,trim=1.6cm 3mm 3.6cm 1cm, height=3cm]{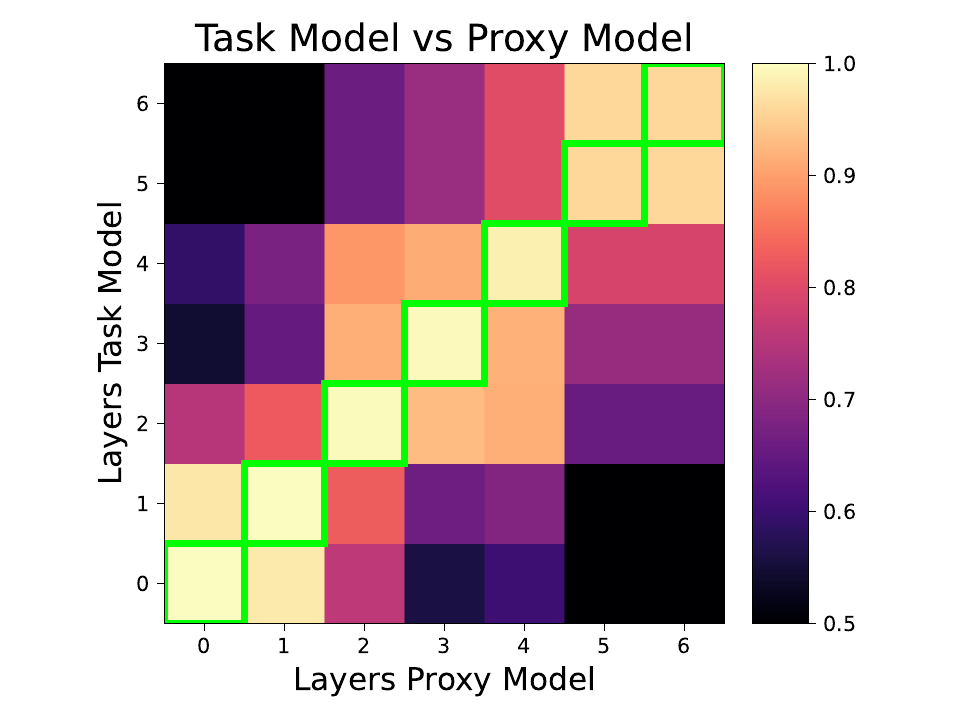}
         \caption{Epoch 30}
     \end{subfigure}
     \hfil
     \begin{subfigure}[b]{0.23\textwidth}
         \centering
         \includegraphics[clip,trim=1.6cm 3mm 3.6cm 1cm, height=3cm]{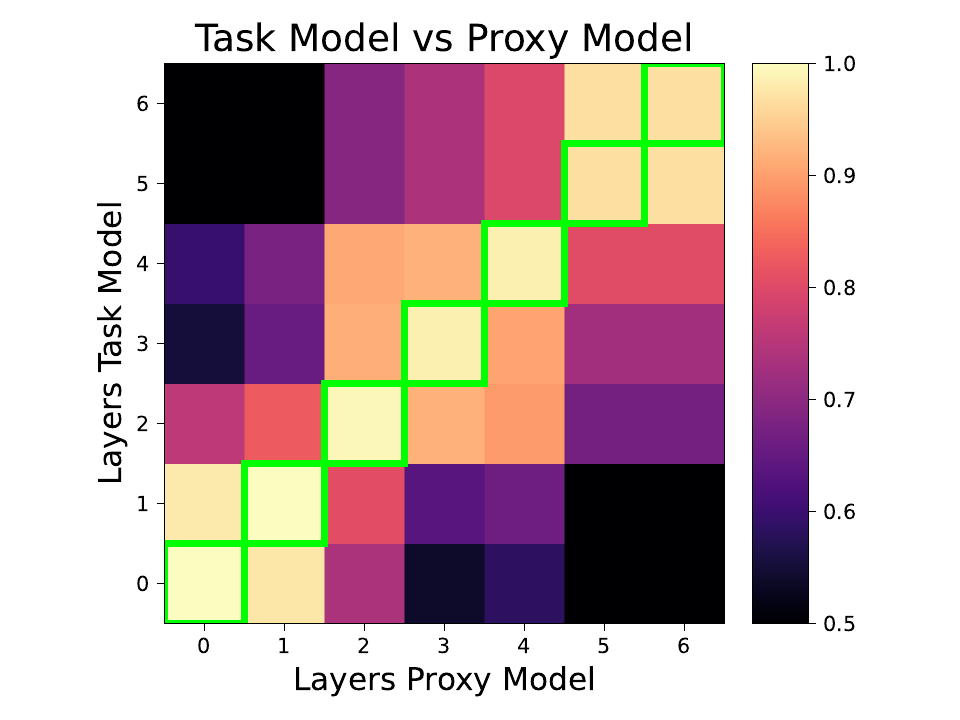}
         \caption{Epoch 60}
     \end{subfigure}
     \hfil
     \begin{subfigure}[b]{0.23\textwidth}
         \centering
         \includegraphics[clip,trim=1.6cm 3mm 3.6cm 1cm, height=3cm]{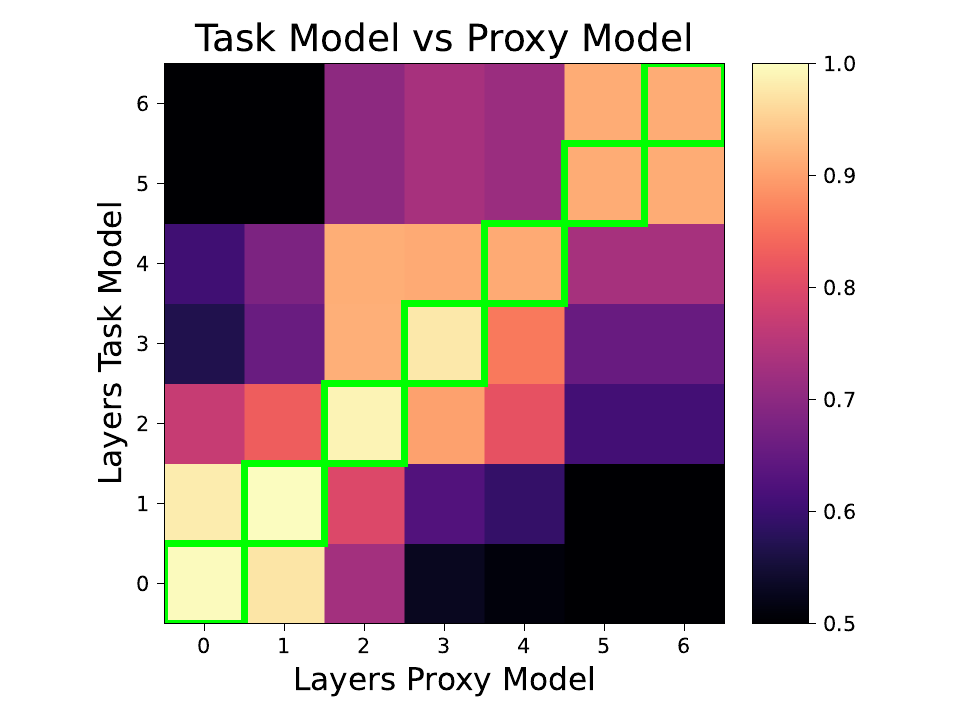}
         \caption{Epoch 90}
     \end{subfigure}
     \hfil    
     \begin{subfigure}[b]{0.25\textwidth}
         \centering
         \includegraphics[clip,trim=1.6cm 3mm 1.6cm 1cm, height=3cm]{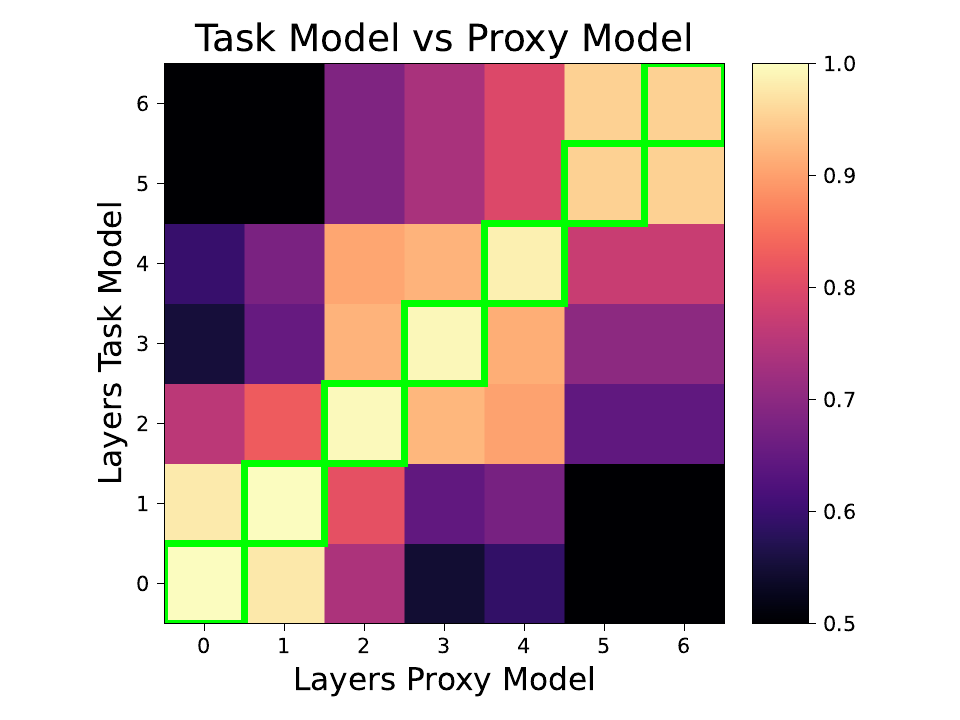}
         \caption{Epoch 120}
     \end{subfigure}
     \hfill
        \caption{Layer-wise feature similarity between the fully updated task model and the proxy model at different epochs. The task model gradually accumulates the knowledge of the proxy model.} 

        \label{fig:cka_peer_task}
\end{figure*}

\begin{table*}[t!]
\caption{The target domain accuracy of the parameter-space ensemble ($^\dagger$ indicates numbers are from original authors).}
\label{tab:ensemble_w_wo}
\centering
\begin{adjustbox}{width=0.8\textwidth}
\centering\begin{tabular}{@{}lcccccccccc@{}}
\toprule
& & \multicolumn{4}{c}{{PACS}} & \multicolumn{5}{c}{{Digits}}\\
\cmidrule(lr){3-6} \cmidrule(lr){7-11}
{Method} & Ensemble & A & C & S & Avg. & SVHN & M-M & S-D & USPS & Avg.\\
\midrule
ERM [\citenum{ermkolt}]& \ding{55}& 54.43 & 42.74 & 42.02 & 46.39 & 27.83 & 52.72 & 39.65 & 76.94 & 49.29\\
MetaCNN$^\dagger$ [\citenum{wan2022}]& \ding{55}& 54.05 & \textbf{53.58} & 63.88 & 57.17 & 66.50 & \textbf{88.27} & 70.66 & 89.64 & 78.76\\
\midrule
P-ENS w/o \textsc{peer}& \ding{51}& \textbf{63.20} & 41.08 & 56.25 & 53.51 & \textbf{71.87} & 76.42 & 82.36 & 92.23 & 80.72 \\
P-ENS \textsc{peer} (ours)& \ding{51} & 62.66 & 47.40 & \textbf{68.21} & \textbf{59.42} & 70.79 & 76.84 &  \textbf{83.05} &  \textbf{93.57} &  \textbf{81.06}\\
\bottomrule
\end{tabular}

\end{adjustbox}
\end{table*}

\subsubsection{Effect of PEER on Learned Features}\label{sec:analysis_features}
In this section, we analyze the \textsc{peer}'s effect on the learned feature representations. In detail, we share two results: (1) parameter-averaging allows the task model to accumulate the proxy model's knowledge, (2) the \textsc{peer} regularization addresses the proxy model's feature distortion (\cref{appendix:effect_on_learned_features}). 

To show this, we follow the practice of \citet{neyshabur2020being} and compute the Centered Kernel Alignment (CKA) metric \citep{kornblith2019similarity} between trained models. The CKA metric measures the similarity between feature representations, where $1.0$ indicates perfect alignment. Specifically, we compute and visualize the CKA similarity for different layers of the multi-layer CNN network trained on the Digits setting (see \cref{appendix:architecture} for details). Each matrix in \cref{fig:cka_peer_task,fig:cka_full} displays the similarity between the two models, its diagonal values indicating the similarity between corresponding layers' features, i.e. brighter boxes indicate more shared knowledge.

We report that the parameter-averaging allows the task model to function similarly to a buffer which accumulates the knowledge of the proxy model across previous training steps. \cref{fig:cka_peer_task}, we illustrate the feature similarity between the task model $F$ ($\theta_{f}$) and the proxy model $P$ ($\theta_{p}$). We can see that the fully updated task model is closely aligned with different stages of the proxy model's trajectory (indicated by bright diagonal values in \cref{fig:cka_peer_task}), suggesting that the parameter-averaging effectively consolidates knowledge from various augmentations and preserves features that might otherwise be distorted during training. Continuing this discussion, on \cref{appendix:effect_on_learned_features}, we show that \textsc{peer} plays an important role in addressing the feature distortion during training (\cref{fig:cka_full}).

\vspace{-1mm}

\subsection{Ablation Study}

We conduct an ablation study to evaluate the impact of various components on overall performance, including the regularization objective (\cref{tab:table_infonce_bt}), 
hyperparameters $w, \lambda$, and $k$ (\cref{tab:ablation_PACS,tab:ablation_PACS_fluctuation})
, model size (\cref{tab:accuracy_modelsize,tab:fluctuation_modelsize})
, and the role of the projection head (\cref{tab:accuracy_projection_head}).

\section{Conclusion}
This paper presents \textsc{peer}, a novel generalization method to address the issues of augmentation-based approaches to single source domain generalization. We highlight the feature distortion induced by augmentation, which triggers fluctuations in the target domain performance during training. Based on our observations, we propose a parameter-averaged task model that accumulates the generalization effect of the training proxy model. Entropy regularization on their learned feature representation aligns the two models, addressing feature distortion. Experiments on various datasets (PACS, Digits, Office-Home, VLCS) demonstrate the effectiveness of our method in stabilizing the learning process and enhancing the generalization performance.

\newpage
\section*{Acknowledgment}
We thank anonymous reviewers for constructive comments to improve the manuscript. This work was partly supported by the IITP (RS-2022-II220953/25\%) and NRF (RS-2023-00211904/50\%, RS-2023-00222663/25\%) grant funded by the Korean government. This work was supported in part through the NYU IT High-Performance Computing resources, services, and staff expertise.

{
    \small
    \bibliographystyle{ieeenat_fullname}
    \bibliography{main}
}

\appendix

\newpage
\begin{figure*}[tb]
     \centering
     \hfil
     \begin{subfigure}[b]{0.44\linewidth}
         \centering
         \includegraphics[width=\linewidth]{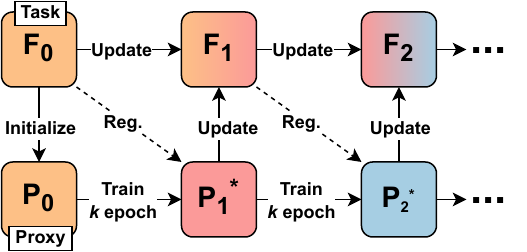}
         \caption{\textsc{peer} Framework}
         \label{fig:peer-peer}
     \end{subfigure}
     \hfil
     \begin{subfigure}[b]{0.44\linewidth}
         \centering
         \includegraphics[width=\linewidth]{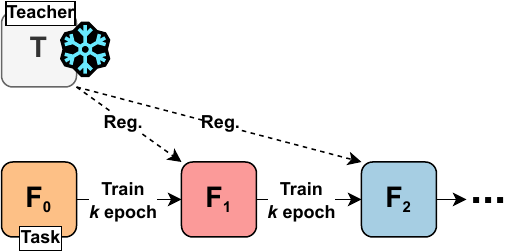}
         \caption{Conventional Teacher-Student Framework}
         \label{fig:teacher-student}
     \end{subfigure}
     \hfil\null
     \caption{The \textsc{peer} framework consists of two interacting modules: a proxy model $P$ and the task model $F$. During training, the task model retains the knowledge of the proxy model via parameter-averaging. The conventional teacher-student framework consists of a frozen teacher $T$ and the task model $F$. Unlike \textsc{peer}, the teacher is not updated, posing limitations in improving task model generalization.}
     \label{fig:framework}
\end{figure*}    

\section{Study on Model-to-Model Regularization}\label{appendix:model-to-model}

In this section, we further study the topic of model-to-model regularization. We first begin by revisiting previous works on model-to-model regularization, highlighting the differences from our approach. Next, we provide experimental results on using a pre-trained teacher for regularization (i.e., teacher-student regularization). Using this, we show the strength of our approach against previous model-to-model regularization methods.

\paragraph{Previous Methods: teacher-student regularization.}
Model-to-model regularization is frequently used to boost a model's performance in tasks such as knowledge distillation \citep{hinton2015distilling,beyer2022knowledge} or generalization \citep{MIRO,li2023simple}. Here, an underlying idea is that the supervisor (i.e. teacher) be a model displaying strong performance, namely OOD robustness. A common approach is to use a pre-trained model trained on a large dataset, or with a larger model architecture. Please refer to \cref{fig:teacher-student} for a better understanding of this approach. However, issues exist in deploying strong teacher models for the sDG task. First, using pre-trained teacher models contradicts the grounding idea of single source domain generalization (sDG). To our understanding, the goal of sDG is to devise a generalization method that can function well in a realistic environment where the source data is limited. Reflecting this, the sDG setting strictly forbids the use of additional source domains for training. In this sense, using a model that is already trained on a much larger dataset seems to go against this. Furthermore, if the teacher model is available for use, a more efficient method would be to directly utilize the teacher for inference, while its operating cost would be much larger.

\paragraph{Our Method: Using a group of \textsc{peer} for regularization.}
Our approach to model-to-model regularization alleviates the irony of using a pre-trained teacher model by replacing it with a parameter-space ensemble (task model $F$). Unlike previous approaches \citep{MIRO,li2023simple}, the \textsc{peer} does not violate the constraints of the sDG setting. Specifically, the task model in \textsc{peer} does not use additional training data as it is the training model itself. Second, it is of an identical architecture to the training proxy model, hence we need not worry about excessive computation costs. Furthermore, using a task model regulator of the identical architecture allows the proxy model to directly update the task model via parameter-averaging, without additional cost. On the other hand, when using a pre-trained teacher model, updating the teacher would require excessive costs (e.g., online distillation \citep{Gou_2021}). 

More importantly, our approach to model-to-model regularization is more easily applicable to real-world problems than using a pre-trained teacher, owing to the adaptive nature of the task model. In \textsc{peer}, the task model is created during the training process. Hence, the task model effortlessly adapts to the new dataset. This adaptivity makes \textsc{peer} applicable to any given task or dataset. On the other hand, a teacher is a fixed model that is supposedly pre-trained on large datasets. The fixed nature of the teacher limits its applicability, as the teacher would only work if the teacher's pre-trained data is similar to the new training data. For instance, a strong digit classification \citep{deng2012mnist} model will not function well as a teacher for other classification tasks \citep{li2023simple}.

\paragraph{Experiment: \textsc{peer} vs. Teacher}

In this section, provide detailed information on our experimental results reported in \Cref{sec:exp_m2m}, and emphasize the competitiveness of \textsc{peer} against using a strong teacher model for regularization. Specifically, we demonstrate that a task model in \textsc{peer} serves as a more robust regulator compared to a pre-trained teacher model. Specifically, we empirically show that a suitable teacher model is not always available. For analysis, we use the PACS and Digits datasets and compare three model-to-model regularization methods (1) None: The baseline without model-to-model regularization (2) Teacher: Following the practice of \citet{MIRO}, we selected the pre-trained RegNetY-16GF \citep{radosavovic2020designing} as a teacher for PACS. In contrast, in Digits, we could not obtain a pre-trained model fit for use as the teacher. Hence, we follow the practice of \citet{MIRO} and use a model pre-trained on both the source and target domains of Digits. We will later elaborate on why the RegNetY-16GF does not apply to the Digits experiment. (3) \textsc{peer}: The task model in \textsc{peer} has the same architecture as the proxy model. At the beginning of training, it is identical to the proxy model and then updated during the training process by averaging the parameters of the proxy model and the task model. The model is trained with random augmentation and follows the setup stated in \Cref{sec:experiment}.

We share the results of the experiment in \Cref{tab:peer_vs_teacher}. Here, the methods $\textsc{t+ra}$ and $\textsc{p+ra}$ refer to applying the teacher regularization and the \textsc{peer} regularization, respectively. First, we compare the effectiveness of the two regulators (the teacher and the task model in \textsc{peer}) in reducing the OOD target domain performance fluctuation. In \Cref{tab:peer_vs_teacher}, we see that both the teacher and the task model in \textsc{peer} reduce the OOD fluctuation (measured as variance), while the teacher displays a stronger regularization effect than the task model. We view that this result reflects the reality that the teacher is a fully trained model, while the task model is updated alongside the proxy model's training process, and hence is a weak supervisor, at least at the beginning of training \citep{burns2023weaktostrong}. On the other hand, we see that the \textsc{peer} shows higher sDG target domain accuracy ($59.42$) in PACS than using a teacher ($56.50$). We believe that this results from the nature of the frozen teacher. To illustrate, the teacher is a frozen model, and hence a model regularized by the teacher may have been bound by the teacher's supervision. On the other hand, the \textsc{peer} uses a task model that grows alongside the proxy model, and hence less likely to share the issues exhibited by the teacher. This pattern is repeated in the Digits experiment at \Cref{tab:peer_vs_teacher}, where the teacher was slightly better in reducing the fluctuation, while our method with \textsc{peer} showed a higher target domain accuracy. 

In \Cref{tab:peer_vs_teacher}, we also test the case when the task model is not updated with parameter-averaging i.e., \textsc{peer} (w/o ParamAvg.). Instead of updating the task model via parameter-averaging, we simply froze a snapshot of the proxy model every $k$ epoch and used it as the regulator. Here, we can see that the non-averaged task model showed effectiveness in alleviating the OOD fluctuation while limiting the target domain accuracy. 

\begin{tcolorbox}[colframe=black, colback=blue!2!, coltitle=black, width=\linewidth, boxrule=0.5mm]
    \textbf{Takeaway:} Model-to-model regularization, regardless of the type of the regulator, can reduce the OOD fluctuation amidst training. However, updating the regulator (task model) is critical to enhancing the target domain accuracy.
\end{tcolorbox}

We find that for certain tasks, a teacher model is hard to obtain. In other words, there is no universal model for use as the teacher. For instance, in the PACS experiment, the RegNetY-16GF displayed sufficient capabilities as a model-to-model regularize. However, using the RegNetY-16GF as the teacher for the Digits experiment was not available. Notably, RegNetY-16GF marked low validation accuracy in the target domain, nor was it able to guide the proxy model. We believe that this difference is derived from the discrepancy between the two datasets. For instance, PACS is a collection of images without any distortion, while Digits is a dataset solely comprised of digit images. Hence, we view that the large gap between the pre-trained dataset of the RegNetY-16GF and the Digit classification datasets is responsible for this behavior. This issue can be explained with the work of \citet{nft}, where the authors demonstrate that there exists a trade-off between a model's performance on a certain task and the performance on all remaining tasks. In contrast, the \textsc{peer} applies to any task, as it gradually adapts to the dataset using the proxy model.

\section{Discussions}

\subsection{Discussion on the fluctuation}

We illustrate the mid-train OOD fluctuation in \Cref{fig:what_is_fluctuation}. Here, the worst-case performance of the fluctuating model (blue) consistently falls below that of the stable model (orange). This describes the issues of deploying a fluctuating model, as the fluctuation poses challenges in early stopping and model selection. 

\citet{arpit2022ensemble} has studied a similar phenomenon within the multi-DG literature, attributing the fluctuation to the stochastic nature of the learning process (e.g., random seed, order of data). 
While we acknowledge the role of other contributing factors, we hypothesize that the mid-train OOD fluctuation primarily stems from the model's inability to accumulate the knowledge learned from varying augmentations. In specific, we view that the model's trained features are distorted, or forgotten during training \citep{kumar2022finetuning,shi2024unified}.

\subsection{Discussion on \textsc{peer} as a Mutual Information Optimization}\label{appendix:mi_optimization}

Here, we further elaborate on the \textsc{peer}. Specifically, we elaborate on why optimizing with \textsc{peer} can maximize the mutual information (MI). To recapitulate, the \textsc{peer} aims to maximize the MI between the output feature representations of the task model $F$ and the proxy model $P$. However, directly optimizing MI is challenging, as its exact estimation is intractable \citep{paninski2003}.  The InfoNCE loss \citep{oord2018} adopts a lower bound of MI  \citep{poole2019variational} as a surrogate objective for MI optimization:
\begin{align} 
I(z;z^+) \geq \Tilde{I}_{\textsc{INCE}}(z; z^+) = -\log \frac{\exp\left(\operatorname{sim}(z, z^+) \right)}{\sum_{k=1}^N \exp\left(\operatorname{sim}(z, z_k) \right)},
\label{loss:infonce}
\end{align}

where $z,z^+$ denotes the feature representations of the original sample $x$ and its augmented view $\bar{x}$, and $\operatorname{sim}$ a similarity function, such as cosine similarity or dot product. The actual computation involves an empirical estimation between a batch of representations of size $N$.

However, an issue of InfoNCE as a variational bound of MI is that InfoNCE requires a large batch size for convergence \citep{shrivastava2021estimating,hjelm2019learning}, making it doubtful for use in small datasets (e.g., PACS). Consequently, in our implementation, we approximate InfoNCE with the feature decorrelation loss \Cref{loss:bt}, based on empirical and theoretical results that show its functional proximity \citep{huang2021,tao_siamese}. Contrary to InfoNCE, the feature decorrelation converges effectively with small batch sizes and large vector dimensions, fit for many sDG settings with smaller datasets, or with images of large sizes.

BT (Barlow Twins), is a feature decorrelation loss \citep{barlow_twins}:

\begin{align}
\operatorname{BT}(Z,Z^+) = \sum_{i}(1-M_{ii})^2 \ + \lambda \sum_{i}\sum_{j \neq i} M_{ij}^2, \label{loss:bt} 
\end{align}

where $M$ refers to the empirical cross-correlation matrix of the two batches of feature representations $Z$, $Z^+$, and $\lambda$ is a balancing coefficient. 
The first term $\sum_{i}(1-M_{ii})^2$ aligns two representations by spurring the diagonal values in $M$ of $(Z,Z^+)$  to be $1$.
The second term $\sum_{i}\sum_{j \neq i} M_{ij}^2$ minimizes redundancy in the representation by encouraging the off-diagonal values to be closer to $0$.

\begin{table*}[t!]
\caption{Target domain accuracy with different entropy regularization functions.}
\label{tab:table_infonce_bt}
\centering
\begin{adjustbox}{width=0.77\textwidth}
\centering\begin{tabular}{@{}llccccccccc@{}}
\toprule
& & \multicolumn{4}{c}{{PACS}} & \multicolumn{5}{c}{{Digits}}\\
\cmidrule(lr){3-6} \cmidrule(lr){7-11}
{Method} & Reg. Obj. & A & C & S & Avg. & SVHN & M-M & S-D & USPS & Avg.\\
\midrule


PEER (ours) & BT [\citenum{barlow_twins}] & \textbf{62.66} & 47.40 & \textbf{68.21} & \textbf{59.42} & \textbf{70.79} & \textbf{76.84} &  \textbf{83.05} &  93.57 &  \textbf{81.06}\\
PEER (ours) & InfoNCE [\citenum{oord2018}] & 60.03 & \textbf{48.11} & 67.91 & 58.68 & 68.34 & 75.80 &  82.69 &  \textbf{93.92} & 80.19\\
\bottomrule
\end{tabular}

\end{adjustbox}
\end{table*}

In \Cref{tab:table_infonce_bt}, we report the experimental results of replacing our regularization objective \cref{loss:bt} with the InfoNCE. We find that both objectives are effective, while our default objective showed stronger results. We believe there are several factors behind this result (e.g., batch size, dataset \citep{balestriero2023cookbookselfsupervisedlearning}).

\section{Effect of \textsc{peer} on the model}

In this section, we further analyze the effect of \textsc{peer}, namely on the proxy model's learned features and its loss landscape.

\subsection{Effect on Learned Features (continued)}\label{appendix:effect_on_learned_features}

\begin{figure*}[tb]
     \centering
     \begin{subfigure}[b]{0.23\textwidth}
         \centering
         \includegraphics[clip,trim=1.6cm 3mm 3.6cm 1cm, height=3cm]{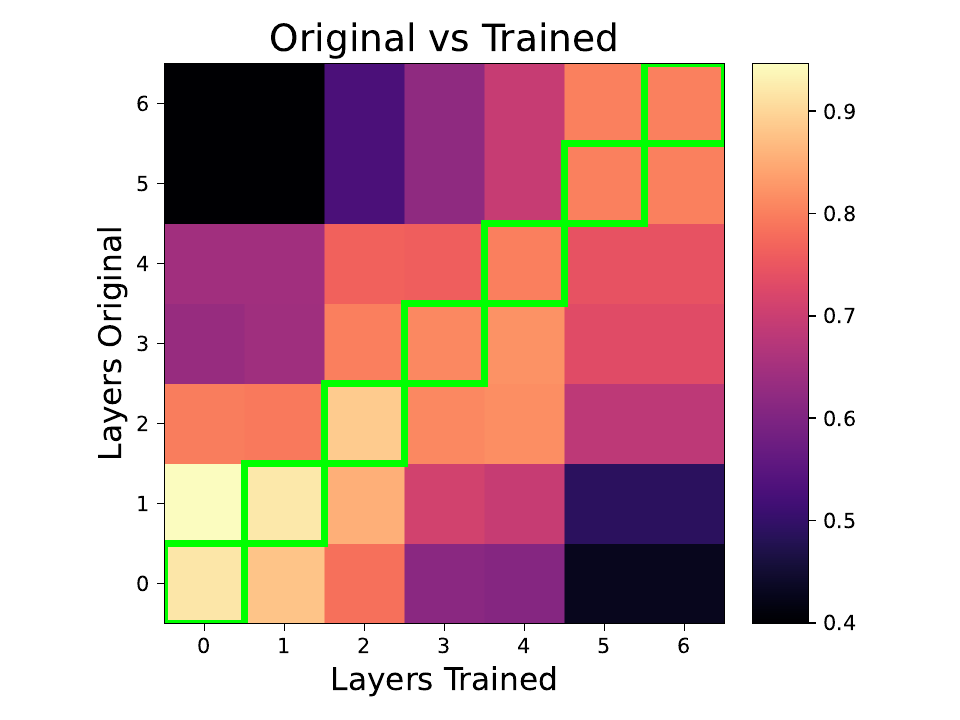}
         \caption{Epoch 30 (w/o \textsc{peer})}
         \label{fig:cka_erm_30}
     \end{subfigure}
     \hfill
     \begin{subfigure}[b]{0.23\textwidth}
         \centering
         \includegraphics[clip,trim=1.6cm 3mm 3.6cm 1cm, height=3cm]{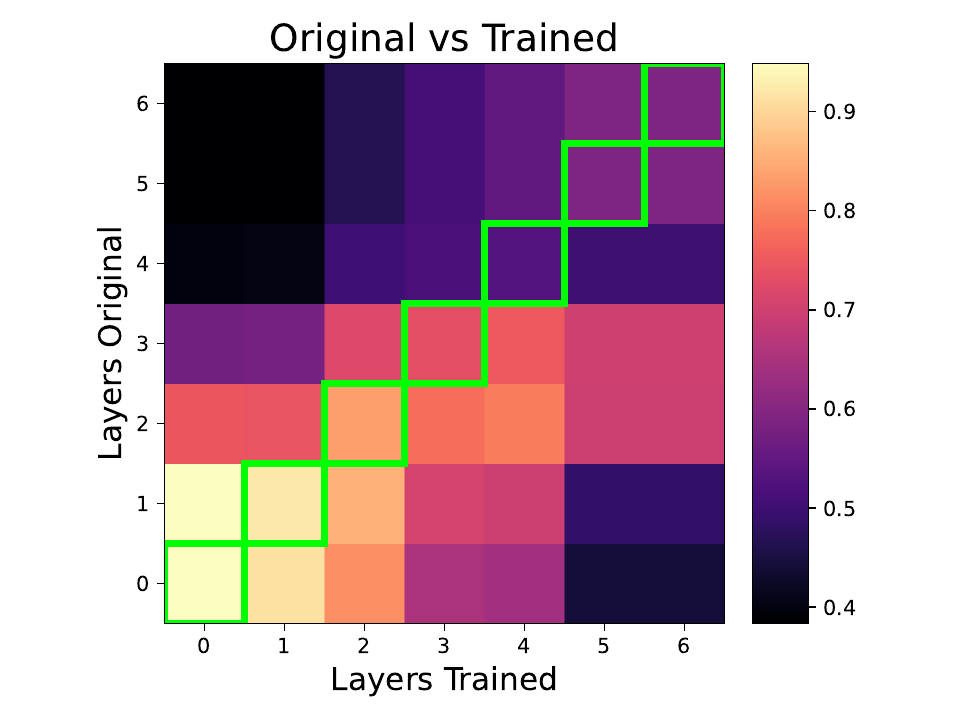}
         \caption{Epoch 120 (w/o \textsc{peer})}
         \label{fig:cka_erm_120}
     \end{subfigure}
     \hfill
     \begin{subfigure}[b]{0.23\textwidth}
         \centering
         \includegraphics[clip,trim=1.6cm 3mm 3.6cm 1cm, height=3cm]{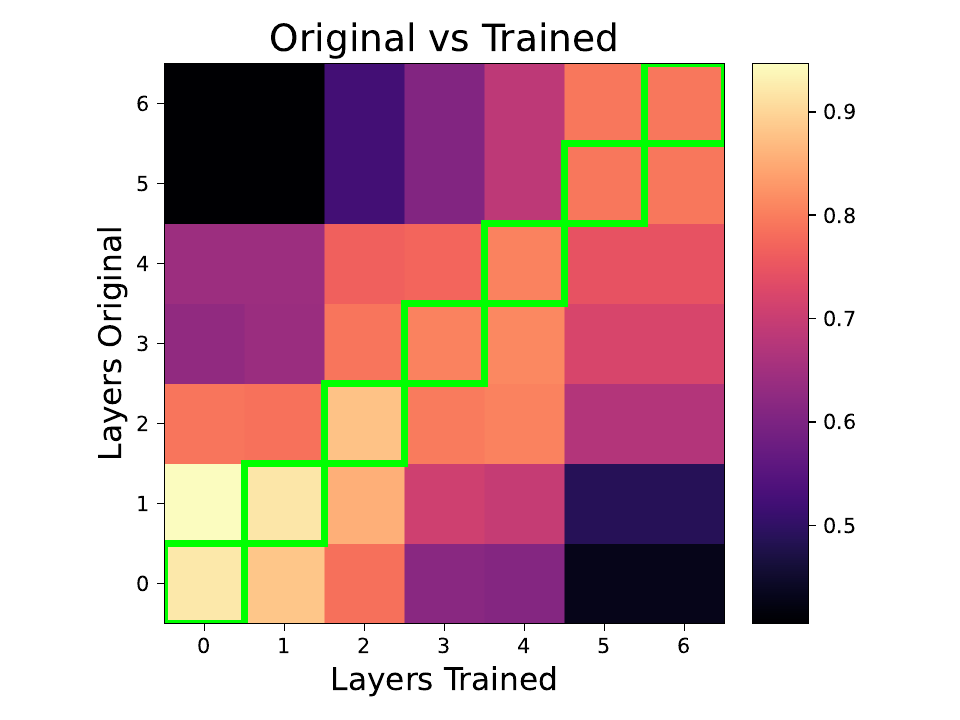}
         \caption{Epoch 30 (w/ \textsc{peer})}
         \label{fig:cka_peer_30}
     \end{subfigure}
     \hfill     
     \begin{subfigure}[b]{0.25\textwidth}
         \centering
         \includegraphics[clip,trim=1.6cm 3mm 1.6cm 1cm, height=3cm]{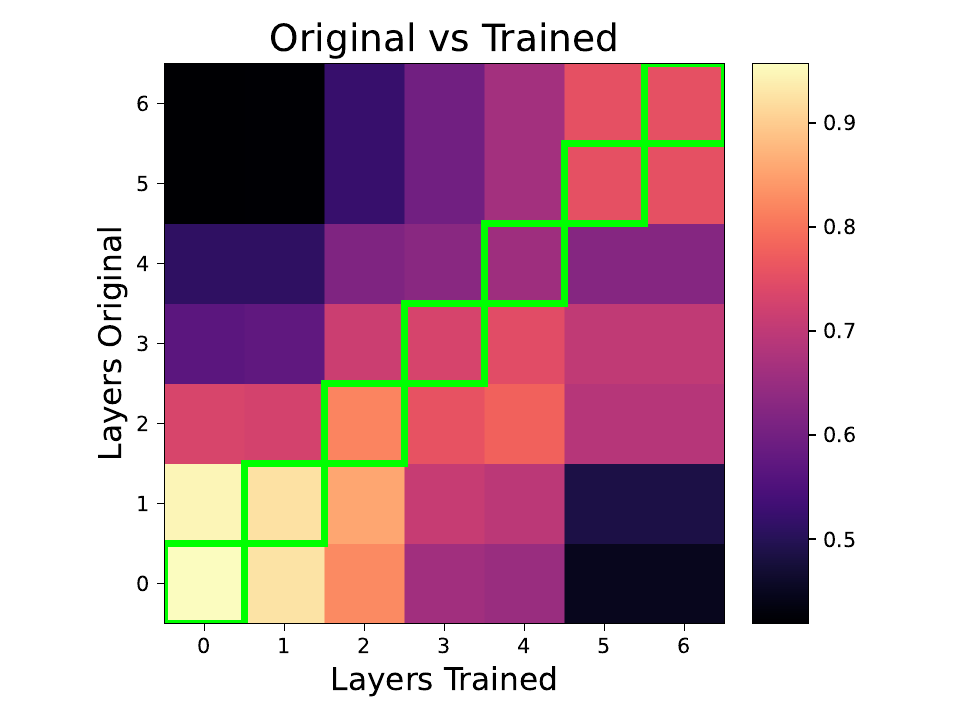}
         \caption{Epoch 120 (w/ \textsc{peer})}
         \label{fig:cka_peer_120}
     \end{subfigure}
        \caption{Layer-wise feature similarity (CKA) between the proxy model after initialization and after training with different epochs. Without \textsc{peer} regularization, the model suffers feature distortion.}
        \label{fig:cka_full}
\end{figure*}

In this section, we study the effect of \textsc{peer} on the learned feature representations. We show that regularization plays an important role in reducing the proxy model's feature distortion during training. We compare two cases (a) \textit{Without} \textsc{peer}: CKA similarity of the proxy model $P$ at different epochs of training and its original state before training (b) \textit{With} \textsc{peer}: CKA similarity of the \textsc{peer} applied proxy model $P$ at different epochs $n$ ($\theta_{p}^{(n)}$) and its original state ($\theta_{p}^{(0)}$). Notably, the diagonal elements in \Cref{fig:cka_peer_120} are brighter in color than their counterparts (\Cref{fig:cka_erm_120}), which indicates that \textsc{peer} allows the proxy model to preserve its pre-trained features. The model is trained with random augmented MNIST data, and the feature similarity is also computed on the MNIST data.

\begin{figure*}[tb]
     \centering
     \hfill
     \begin{subfigure}[b]{0.24\textwidth}
         \centering
         \includegraphics[clip,trim=1.6cm 3mm 3.6cm 1cm, height=3cm]{images/cka/cka_erm_30.pdf}
         \caption{Epoch 30}
     \end{subfigure}
     \hfill
     \begin{subfigure}[b]{0.24\textwidth}
         \centering
         \includegraphics[clip,trim=1.6cm 3mm 3.6cm 1cm, height=3cm]{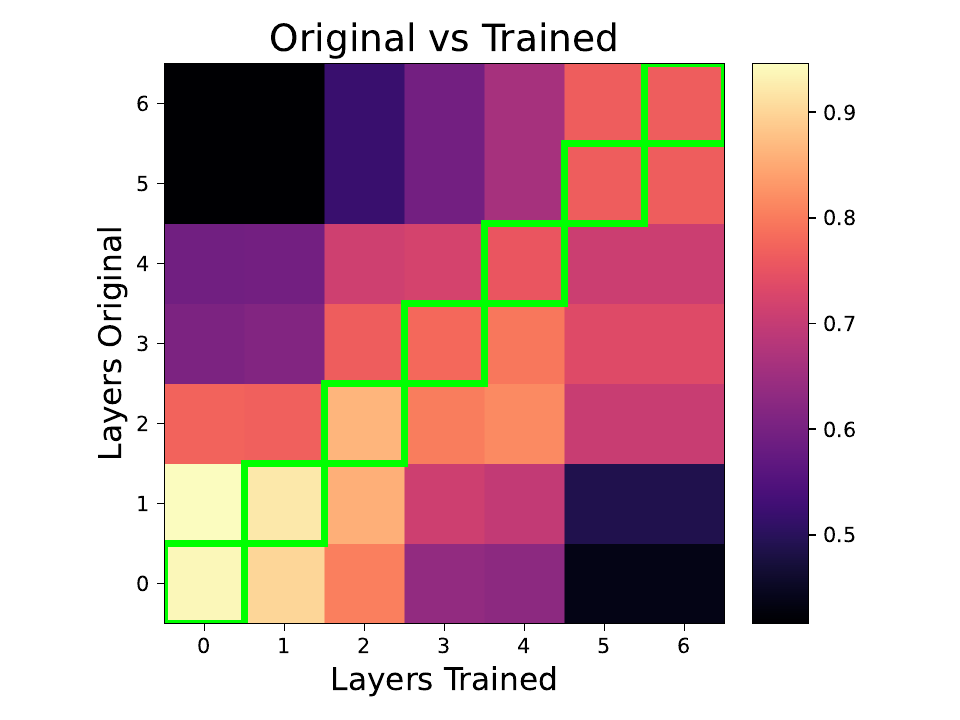}
         \caption{Epoch 60}
     \end{subfigure}
     \hfill
     \begin{subfigure}[b]{0.24\textwidth}
         \centering
         \includegraphics[clip,trim=1.6cm 3mm 3.6cm 1cm, height=3cm]{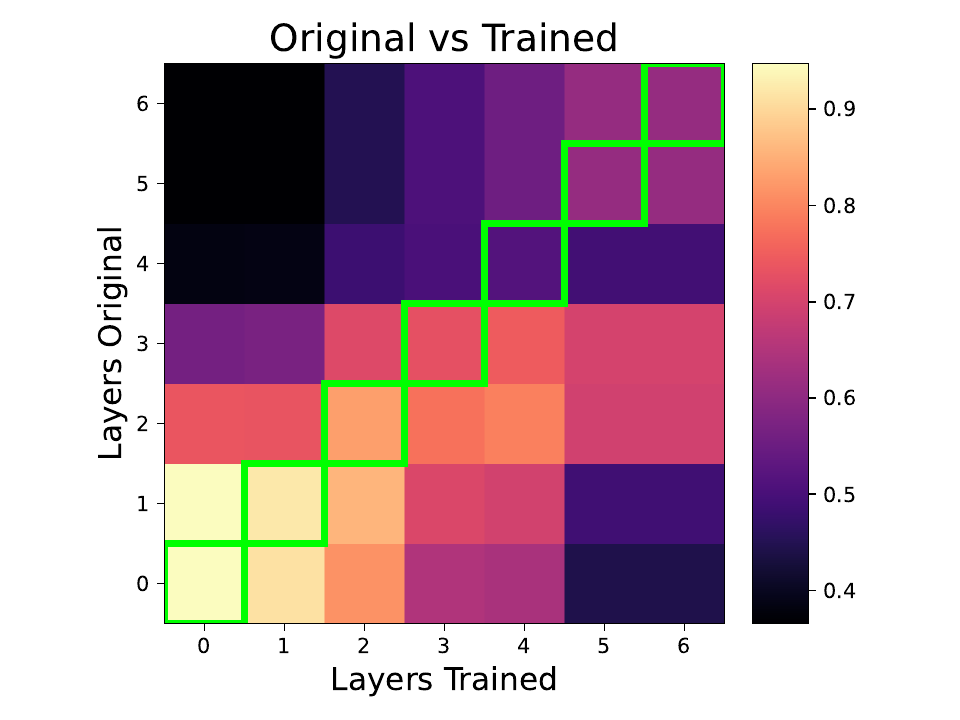}
         \caption{Epoch 90}
     \end{subfigure}
     \hfill     
     \begin{subfigure}[b]{0.24\textwidth}
         \centering
         \includegraphics[clip,trim=1.6cm 3mm 1.6cm 1cm, height=3cm]{images/cka/cka_erm_120.pdf}
         \caption{Epoch 120}
     \end{subfigure}
     \hfill\null
        \caption{Layer-wise Feature Similarity (CKA) between the proxy model's initialization and the trained proxy model (without \textsc{peer}). Without \textsc{peer} regularization, the model suffers feature distortion.}
        \label{fig:cka_erm}
\end{figure*}

\begin{figure*}[tb]
     \centering
     \hfill
     \begin{subfigure}[b]{0.24\textwidth}
         \centering
         \includegraphics[clip,trim=1.6cm 3mm 3.6cm 1cm, height=3cm]{images/cka/cka_peer_30.pdf}
         \caption{Epoch 30}
     \end{subfigure}
     \hfill
     \begin{subfigure}[b]{0.24\textwidth}
         \centering
         \includegraphics[clip,trim=1.6cm 3mm 3.6cm 1cm, height=3cm]{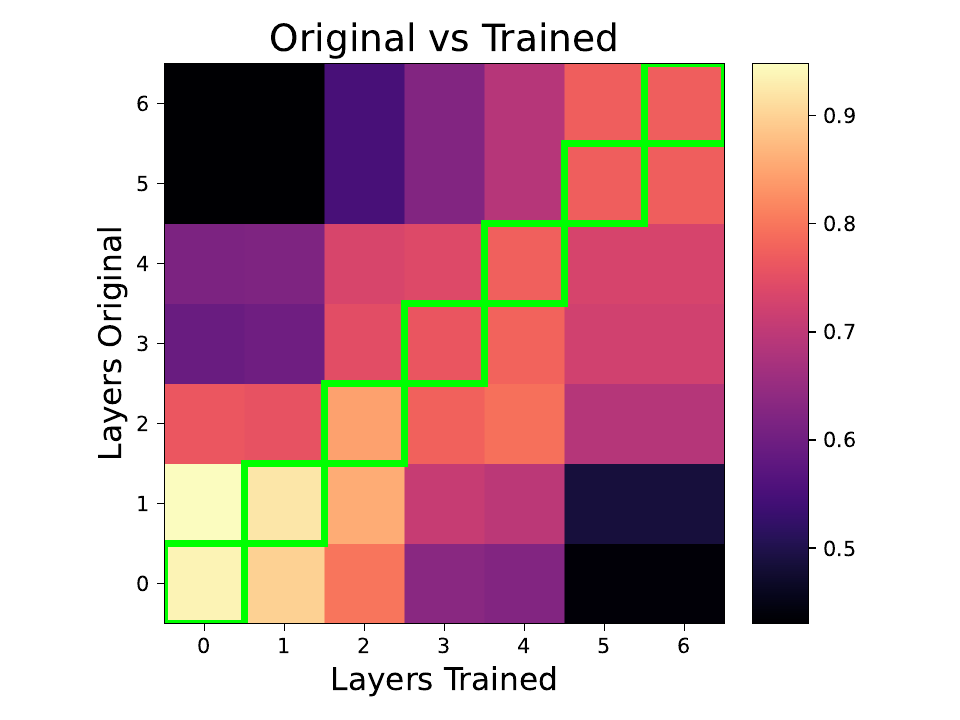}
         \caption{Epoch 60}
     \end{subfigure}
     \hfill
     \begin{subfigure}[b]{0.24\textwidth}
         \centering
         \includegraphics[clip,trim=1.6cm 3mm 3.6cm 1cm, height=3cm]{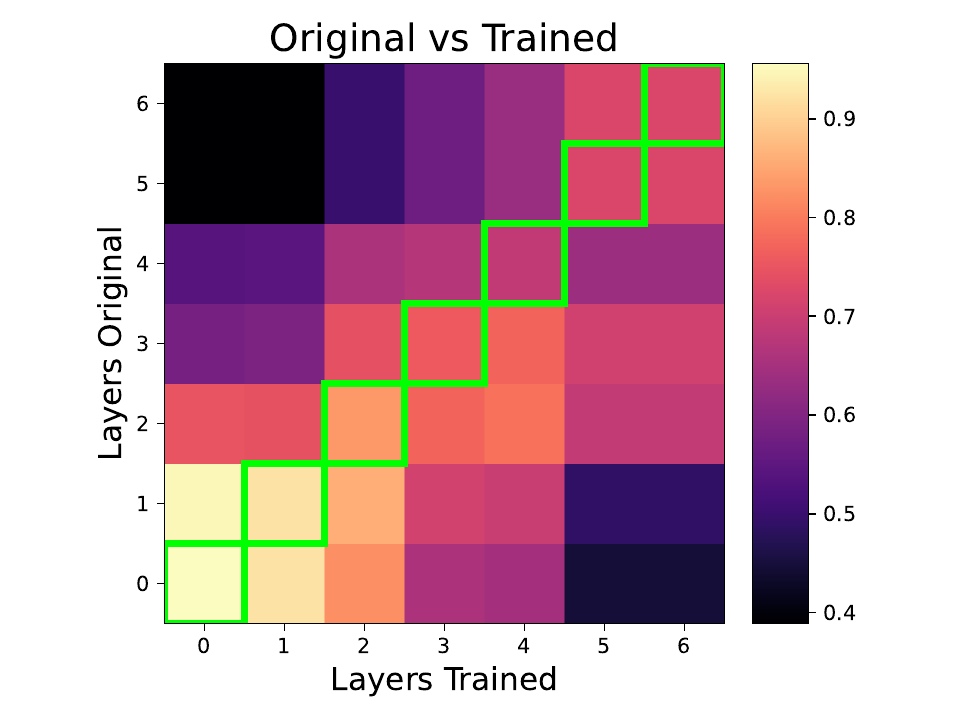}
         \caption{Epoch 90}
     \end{subfigure}
     \hfill     
     \begin{subfigure}[b]{0.24\textwidth}
         \centering
         \includegraphics[clip,trim=1.6cm 3mm 1.6cm 1cm, height=3cm]{images/cka/cka_peer_120.pdf}
         \caption{Epoch 120}
     \end{subfigure}
     \hfill\null
         \caption{Layer-wise Feature Similarity (CKA) between the proxy model's initialization and the trained proxy model (with \textsc{peer}). With \textsc{peer}, the model suffers less feature distortion.}
        \label{fig:cka_peer}
\end{figure*}

Next, we provide a more detailed analysis. In \Cref{fig:cka_erm}, we report the case where there is no regularization from the task model (without \textsc{peer}). Here, the diagonal values indicate the corresponding layers between the initialization and the trained model. We can see that as training continues (\Cref{fig:cka_erm_120}), a lot of trained knowledge is distorted in the later layers of the model. In contrast, \Cref{fig:cka_peer} shows that when regularized with the task model (with \textsc{peer}), the proxy model preserves a lot of knowledge even in the later epochs (\Cref{fig:cka_peer_120}). Yet, we do not claim that \textsc{peer} allows the proxy model to perfectly preserve its trained knowledge amidst diverse augmentation \citep{nft}. Rather, we believe that by regularizing the proxy model, we can ultimately benefit the parameter-averaged task model. In the following section, we will empirically show that the regularization indeed benefits the parameter-averaging.

\begin{tcolorbox}[colframe=black, colback=blue!2!, coltitle=black, width=\linewidth, boxrule=0.5mm]
    \textbf{Takeaway:} Model-to-model regularization with \textsc{peer} helps preserve previously learned features in both the task model $F$ and the proxy model $P$.
\end{tcolorbox}

\subsection{Effect on Parameter-Averaging (continued)}\label{appendix:mode_connectivity}

In this section, we provide an extended analysis of how regularizing the proxy model $P$ with the task model (i.e., \textsc{peer}) aids parameter averaging. We argue that the regularization aids the ensembling effect by aligning different snapshots of the proxy model $\theta_p^{(i)}, \theta_p^{(j)}$ that were trained on very different augmented domains.

To show this, we perform a simple experiment: "Can parameter-averaging proxy model snapshots without regularization create a robust regulator?". Similar to \textsc{peer} update, we periodically save snapshots of the proxy model training with random augmentation for every $k$ epoch. The experiment takes place in the PACS and the Digits benchmarks, and follows the same setting stated in \Cref{sec:experiment}. For PACS, the proxy model is trained for $200$ epochs with random augmented data, where $k$ is set as $10$. In Digits, the model is trained for $100$ with $k$ set as $10$. After training, we parameter average the saved snapshots to form a parameter-space ensemble. Note that in this case, no regularization took place. 

We share the results in \Cref{tab:ensemble_w_wo}. As a recap, we explain the notations used in \Cref{tab:ensemble_w_wo}. In the table, P-ENS refers to the parameter-space ensembles. In both PACS and Digits, parameter-space ensembling with regularization (\textsc{peer}) outcompetes ensembling without regularization (P-ENS w/o \textsc{peer}). Notably in PACS, we observe failure cases of parameter-space ensembling without regularization, where the ensemble effect (i.e., gain in generalization ability) was very marginal. As noted in \Cref{sec:effect_paramavg}, this failure case is noteworthy since parameter averaging across different training snapshots of models with the same initialization has been highly successful in many prior studies \citep{grill2020bootstrap,izmailov2018averaging}.

Generally, for a parameter-averaged model to display ensemble effects, some conditions should be simultaneously met \citep{rame2023model}. (1) Share an identical initialization: models that share an initialization backbone tend to display very low loss barriers, showing mode connectivity. (2) Trained on same data: Models trained on identical source data \citep{choshen2022fusing} tend to display mode connectivity, while models trained on varying data commonly do not \citep{ainsworth2023git}. In our case, the first condition is already met, while the second condition may have been broken due to the varying effects of data augmentation. Drawing from this, we hypothesize that the failure case above potentially derives from violating the second condition. In specific, we believe that the discrepancy between two very different augmented domains breaks the alignment between the model snapshots. In this sense, the \textsc{peer} may help parameter-space ensembling by encouraging the regularized proxy model to align the newly augmented domain to the task model's source domain \Cref{sec:peer_regularization}. Unfortunately, the alignment of models in its loss landscape is a topic that has not yet been thoroughly analyzed from a theoretical perspective, especially for models with deep architectures. While our empirical analysis may provide some insight, we believe further research is required on this topic. 

\begin{tcolorbox}[colframe=black, colback=blue!2!, coltitle=black, width=\linewidth, boxrule=0.5mm]
    \textbf{Takeaway:} Model-to-model regularization with \textsc{peer} benefits parameter-averaging between the task model $F$ and the proxy $P$ by aligning the two in the feature space, within a close loss basin.
\end{tcolorbox}

\begin{table*}[t!]
\caption{Ablation study on different components of PEER. Target domain accuracy on PACS and Digits.}
\label{tab:noaug}
\centering
\begin{adjustbox}{width=0.7\textwidth}
\centering\begin{tabular}{@{}lccccccccc@{}}
\toprule
& \multicolumn{4}{c}{{PACS}} & \multicolumn{5}{c}{{Digits}}\\
\cmidrule(lr){2-5} \cmidrule(lr){6-10}
{Method} & A & C & S & Avg. & SVHN & M-M & S-D & USPS & Avg.\\
\midrule
PEER (No Aug.) & 52.46 & 42.02 & 53.35 & 49.28 & 29.19 & 54.14 & 41.06 & 78.33 & 50.68\\
PEER (No ParamAvg.) & 57.73 & 46.69 & 61.33 & 55.25 & 59.99 & \textbf{77.26} & 72.30 & 88.28 & 74.46\\
PEER (No Reg.) & \textbf{63.20} & 41.08 & 56.25& 53.51 & \textbf{71.87} & 76.42 & 82.36& 92.23& 80.72\\
PEER (Ours) & 62.66 & \textbf{47.40} & \textbf{68.21} & \textbf{59.42} & 70.79 & 76.84 &  \textbf{83.05} &  \textbf{93.57} &  \textbf{81.06}\\

\bottomrule
\end{tabular}

\end{adjustbox}
\end{table*} 

\begin{table*}[t!]
\centering
\caption{(a) Target domain accuracy and (b) fluctuation on PACS with different hyperparameters.}
\label{tab:ablation_PACS_combined}
\begin{subtable}[t]{0.48\textwidth}
    \centering
    \caption{Target domain accuracy}
    \label{tab:ablation_PACS}
    \begin{adjustbox}{width=\textwidth}
        \begin{tabular}{@{}llcccc@{}} 
\toprule
{Method} & Hyperparam. & A & C & S & Avg.  \\
\midrule
\multicolumn{6}{c}{Hyperparameter: $w$}\\
\midrule
Ours & w = 0.1 & 59.96 & 45.83 & 66.57 & 57.45 \\
Ours & w = 0.5 & 60.07 & 46.11 & 66.2 & 57.46 \\
Ours & w = 1.0 & 61.22 & 46.20 & 65.79 & 57.74 \\
Ours & w = 2.0 & 61.20 & 46.08 &  66.00 & 57.56 \\
Ours & w = 4.0 & 59.99 & 45.84 & 63.51 & 56.45 \\
Ours & w = 10.0 & 60.14 &45.88 &65.26 & 57.09\\

\midrule
\multicolumn{6}{c}{Hyperparameter: $\lambda$}\\
\midrule
Ours & $\lambda$ = 0.001 & 60.01 & 47.38 & 66.4 & 57.93 \\
Ours & $\lambda$ = 0.005 & 61.20 & 46.08 &  66.00 & 57.56 \\
Ours & $\lambda$ = 0.01 & 60.78 & 48.25 & 65.2 & 58.08 \\
Ours & $\lambda$ = 0.1 & 61.04 & 45.63 & 66.36 & 57.68 \\

\midrule
\multicolumn{6}{c}{Hyperparameter: $k$}\\
\midrule
Ours & $k$ = 1 & 56.99 & 42.30 & 67.25 & 55.51 \\
Ours & $k$ = 5 & 62.17 & 47.42 & 63.52 & 57.70 \\
Ours & $k$ = 10 & 61.20 & 46.08 &  66.00 & 57.76 \\
Ours & $k$ = 20 & 63.45 & 47.11 & 62.23 & 57.60 \\

\bottomrule
\end{tabular}

    \end{adjustbox}
\end{subtable}
\hfill
\begin{subtable}[t]{0.48\textwidth}
    \centering
    \caption{Variance of target domain accuracy}
    \label{tab:ablation_PACS_fluctuation}
    \begin{adjustbox}{width=\textwidth}
        \begin{tabular}{@{}llcccc@{}} 
\toprule
{Method} & Hyperparam. & A & C & S & Avg.  \\
\midrule
\multicolumn{6}{c}{Hyperparameter: $w$}\\
\midrule
Ours & w = 0.1 & \phantom{0}2.19 & \phantom{0}4.38 & \phantom{0}4.45 & \phantom{0}3.67 \\
Ours & w = 0.5 & \phantom{0}2.05 & \phantom{0}3.91 & \phantom{0}4.82 & \phantom{0}3.59 \\
Ours & w = 1.0 & \phantom{0}2.14 & \phantom{0}4.38 & \phantom{0}4.45 & \phantom{0}3.67 \\
Ours & w = 2.0 & \phantom{0}2.01 & \phantom{0}3.98 & \phantom{0}4.77 & \phantom{0}3.59 \\
Ours & w = 4.0 & \phantom{0}2.44 & \phantom{0}3.77 & \phantom{0}4.75 & \phantom{0}3.65 \\
Ours & w = 10.0 & \phantom{0}2.11 & \phantom{0}4.14 & \phantom{0}4.56 & \phantom{0}3.50 \\

\midrule
\multicolumn{6}{c}{Hyperparameter: $\lambda$}\\
\midrule
Ours & $\lambda$ = 0.001 & \phantom{0}2.13 & \phantom{0}3.65 & \phantom{0}5.22 & \phantom{0}3.67 \\
Ours & $\lambda$ = 0.005 & \phantom{0}2.01 & \phantom{0}3.98 & \phantom{0}4.77 & \phantom{0}3.59 \\
Ours & $\lambda$ = 0.01 & \phantom{0}1.99 & \phantom{0}4.04 & \phantom{0}4.71 & \phantom{0}3.58 \\
Ours & $\lambda$ = 0.1 & \phantom{0}2.44 & \phantom{0}4.16 & \phantom{0}4.58 & \phantom{0}3.73\\

\midrule
\multicolumn{6}{c}{Hyperparameter: $k$}\\
\midrule
Ours & $k$ = 1 &  \phantom{0}2.35 & \phantom{0}4.74 & \phantom{0}4.93 & \phantom{0}4.01 \\
Ours & $k$ = 5 &  \phantom{0}2.14& \phantom{0}4.26 & \phantom{0}4.81 & \phantom{0}3.74 \\
Ours & $k$ = 10 & \phantom{0}2.01 & \phantom{0}3.98 & \phantom{0}4.77 & \phantom{0}3.59 \\
Ours & $k$ = 20 & \phantom{0}2.39 & \phantom{0}3.85 & \phantom{0}4.56 & \phantom{0}3.60 \\

\bottomrule
\end{tabular}

    \end{adjustbox}
\end{subtable}
\end{table*}

\section{Ablation Study}

\subsection{Study on Each Component}\label{appendix:ablation_component}

In this section, we share the results of an ablation study on each of the components in PEER. Specifically, we study the role of each component in (1) data augmentation, (2) parameter-averaging of the task model regulator, and (3) regularization by analyzing its effect on the target domain accuracy. The results are reported in \cref{tab:noaug}. The results in \cref{tab:noaug} indicate that all three components are critical in PEER. Especially, it is worth noting that the main source of performance gain in PEER originates from data augmentation, while the other two components (i.e., parameter averaging and regularization) play a significant role in reliably accumulating the effect of data augmentation for robustness.

\subsection{Study of Hyperparameters}\label{appendix:hyperparameter_experiment}
We explore our method's sensitivity to hyperparameters. ($w$): $w$ is the hyperparameter used in \Cref{loss:f}, which functions as the balancing weight of the ERM objective and the regularization objective \Cref{loss:peer}. We find that $w$ does not severely impact the course of training unless set to $0$. We find that during training, the two losses are automatically tuned to match the magnitude of the $w$. (${\lambda}$): $\lambda$ is the hyperparameter used for \textsc{peer} that operates as the balancing weight of the two functions in \Cref{loss:bt}. We begin with the value in the original paper \citep{barlow_twins} with $\lambda= 0.005$, and an alternate value $\frac{1}{r}$ introduced in \citet{tsai2021} where $r$ is the length of a vector in $\mathcal{R}$ (regularization head output space). We observe that our method is resilient to the switch between two candidate values of $\lambda$ although we cannot guarantee they are optimal. (${k}$):  The augmentation reinitialization criteria $k$ is set as $10$ for all experiments to ensure that the proxy model is sufficiently trained before switching the augmentation strategy. We find that switching $k$ with larger numbers causes no problem in training, but setting them too low $k < 2$ poses issues in aligning the proxy model with the task model, undermining the fluctuation stabilization effect.

We share the experimental results of our study on hyperparameters in \Cref{tab:ablation_PACS} and \Cref{tab:ablation_PACS_fluctuation}. As illustrated above, our method \textsc{peer} showed resilience to changes in  $w$ and $\lambda$. Both the target domain accuracy and the OOD fluctuation were insensitive to the change in these two hyperparameters. However, we find that $k$ affects the fluctuation stabilization effect of our method, where setting $k<1$ resulted in a slightly higher variance ($4.01$). This aligns with our expectations, as the proxy and task model may not benefit from the \textsc{peer} regularization in just a single epoch. However, we discover that $k$ influences the stabilization of fluctuations in our method, with $k<2$ leading to a slightly higher variance ($4.01$). This aligns with our expectations, as the proxy and task model may not fully benefit from the \textsc{peer} regularization within a single epoch.

\subsection{Study of Model Validation \& Selection}

Regarding model selection, we report the performance of the final model without early stopping. Following prior works \citep{wan2022}, the hyperparameters were tuned using the \textit{oracle} test dataset, which has shown stability owing to the parameter-averaging process that functions similarly to an ensemble model. Alternatively, we can adopt an alternative validation approach that does not involve the oracle test dataset. For instance, \citet{efthymiadis2024crafting} introduced a novel validation approach that crafts a simulated validation set through data augmentation. 

Reflecting this, we validate the model on two validation sets (1) Source Val. ($S_v$): The validation set of the source domain, (2) Crafted Val. ($C_v$): Crafted Validation set in \citep{efthymiadis2024crafting}. Test: The model is tested on the true target domain. The results are shared in \cref{tab:source-only}. The models were selected with the best validation accuracy. We empirically reconfirm that \textsc{peer} outcompetes the baselines. 
 
Similarly, we can tune our hyperparameters using the source-generated validation set. Results are reported in \cref{tab:source-only-hyperparam}.

\begin{table*}[ht]
\caption{Test Acc. on PACS, the model selected using Validation Set.}
\label{tab:source-only}
\centering
\begin{adjustbox}{width=0.33\textwidth}
\centering\begin{tabular}{@{}lcc@{}} 
\toprule
Method & $S_v$ $\rightarrow$ Test & $C_v$  $\rightarrow$ Test \\
\midrule
ERM & 43.54 & 49.04 \\
RandAugment & 48.81 & 54.90\\
PEER (ours) & 57.52 & 59.29\\
\bottomrule
\end{tabular}

\end{adjustbox}
\end{table*}

\begin{table*}[ht]
\tiny	
\centering
\caption{Test Acc. of our method on PACS with different hyperparameter values.}
\label{tab:source-only-hyperparam}
\begin{subtable}[t]{0.44\textwidth}
\centering
\caption{$w$}
\label{tab:source-only-hyperparam-w}
\begin{adjustbox}{width=\textwidth}
\begin{tabular}{@{}lccc@{}} 
\toprule
{Method} & Hyperparam. & Crafted ($C_v$) & Test \\
\midrule
Ours & $w$ = 0.1 & 78.13 & 57.45 \\
Ours & $w$ = 0.5 & 78.84 & 57.46 \\
Ours & $w$ = 1.0 & 78.50 & 57.74 \\
Ours & $w$ = 2.0 & 78.86 & 57.76 \\
Ours & $w$ = 10.0 & 78.82 & 57.09\\
\bottomrule
\end{tabular}

\end{adjustbox}
\end{subtable}
\hspace{0.05\textwidth} 
\begin{subtable}[t]{0.44\textwidth}
\centering
\caption{$k$}
\label{tab:source-only-hyperparam-k}
\begin{adjustbox}{width=\textwidth}
\begin{tabular}{@{}lccc@{}} 
\toprule
{Method} & Hyperparam. & Crafted ($C_v$) & Test \\
\midrule
Ours & $k$ = 1 & 74.71 & 55.51 \\
Ours & $k$ = 5 & 78.39 & 57.70 \\
Ours & $k$ = 10 & 78.86 & 57.76 \\
Ours & $k$ = 20 & 79.85 & 57.60 \\
Ours & $k$ = 30 & 79.24 & 57.77 \\
\bottomrule
\end{tabular}

\end{adjustbox}
\end{subtable}
\end{table*}

\begin{tcolorbox}[colframe=black, colback=blue!2!, coltitle=black, width=\linewidth, boxrule=0.5mm]
    \textbf{Takeaway:} Parameter-Averaging of different models can benefit from the entropy regularization before the merging process, which functions as an alignment step. We experimentally find that $2$ or more epochs are sufficient for the alignment.
\end{tcolorbox}

\subsection{Study of Model Size}

In this section, we present our findings on the effect of model size on generalization. We observe that larger models/backbones generally improve target domain accuracy. To demonstrate this, we replaced the backbones in three experiments: switching from AlexNet to ResNet-18 for PACS, and from ResNet-18 to ResNet-50 for Office-Home and VLCS. All backbones (AlexNet, ResNet-18, ResNet-50) were pre-trained on the same Imagenet-1k dataset. We found that as the backbone size increased, target domain accuracy improved (\Cref{tab:ablation_PACS}), though mid-train OOD fluctuation (variance of the target domain accuracy) increased slightly (\Cref{tab:ablation_PACS_fluctuation}). However, the gain in accuracy outweighs the rise in variance, suggesting that larger models enhance generalization. We recommend future work to replace default backbones (e.g., AlexNet for PACS, 3-layer MLP for Digits) with larger ones (e.g., ResNets, ViTs).

\begin{tcolorbox}[colframe=black, colback=blue!2!, coltitle=black, width=\linewidth, boxrule=0.5mm]
    \textbf{Takeaway:} Incrementing the model size significantly enhances the generalization capability. However, the fluctuation persists regardless of the increase in model size.
\end{tcolorbox}

\subsection{Additional Experiments}

\begin{table*}[ht]
\vspace{-2pt}
\caption{Target domain Acc. on various benchmark/architectures.} 
\vspace{-5pt}
\centering
\begin{subtable}[t]{0.33\textwidth}
\centering
\caption{Terra Incognita with Resnet-18.}
\label{tab:terra}
\begin{adjustbox}{width=\textwidth}
\begin{tabular}{@{}l|rrr|r@{}} 
\toprule
\textbf{Method} & L38 & L43 & L46 & Avg.  \\
\midrule
ERM  & 22.90 & \textbf{15.85} & 22.91 & 20.55 \\
RandAug.  & 36.41 & 12.80 & 20.41 & 23.21 \\
ADA & 37.33 & 12.94 & 21.20 & 23.82\\
PDEN   & 37.52 & 14.93 & 20.80 & 24.42\\
Ours   & \textbf{38.94} & 15.07 & \textbf{29.09} & \textbf{27.70}\\

\bottomrule
\end{tabular}
\end{adjustbox}
\end{subtable}
\hspace{0.05\textwidth} 
\begin{subtable}[t]{0.33\textwidth}
\centering
\caption{PACS with ViT.}
\label{tab:ablation_PACS_vit}
\begin{adjustbox}{width=\textwidth}
\begin{tabular}{@{}l|rrr|r@{}} 
\toprule
\textbf{Method} & A & C & S & Avg.  \\
\midrule
ERM & 58.42 &  39.25 & 32.27 & 43.31\\
RandAug. & 61.10 & 44.37 & 54.98 & 53.48\\
ADA & 66.21 & 35.70 & 29.77 &43.90\\
PDEN  & 62.96 & 49.87 & 61.87 & 58.23 \\
Ours  & \textbf{75.06} & \textbf{56.78} & \textbf{70.22} & \textbf{67.36} \\

\bottomrule
\end{tabular}
\end{adjustbox}
\end{subtable}
\end{table*}

\paragraph{Additional Benchmarks} 

We have added new experiments on Terra Incognita (\Cref{tab:terra}),
where PEER outperforms baselines by a large margin. Although the gains of data augmentation are relatively small compared to other datasets, PEER outperforms other methods. 

\paragraph{Additional Model Architectures} 

We also test our method on different model architectures (e.g., Vision Transformers). The results are reported in \Cref{tab:ablation_PACS_vit}, using a ViT model (i.e., ViT-B-16) on PACS. Results indicate that PEER works seamlessly on other model architectures, outperforming all baselines.

\section{Implementation Detail}\label{appendix:implementation}

In this section, we report the implementation details of our method.

\subsection{Datasets}\label{appendix:datasets}

Here, we elaborate on the datasets used in our experiments. 

\textbf{{PACS}} \citep{pacs} consists of 4 domains of differing styles (Photo, Art, Cartoon, and Sketch) with 7 classes. In default, we train our model with the Photo domain and evaluate the remaining target domains.  We use the train/test split provided by the original paper \citep{pacs}.

\textbf{Digits} is comprised of 5 different digit classification datasets, MNIST \citep{deng2012mnist}, SVHN \citep{svhn}, MNIST-M \citep{mnistm}, SYNDIGIT \citep{syndigit}, USPS \citep{usps}. In our experiment, we train our model with the first 10,000 samples of the MNIST dataset and assess its generalization accuracy across the remaining four domains. 

\textbf{Office-Home} \citep{officehome} is a common benchmark for DG, but not for sDG. The benchmark consists of 4 datasets (Real-world, Art, Clipart, Product) with differing styles with 65 classes. We train on the Real-world domain and evaluate the remaining domains.

\textbf{VLCS} \citep{fang2013unbiased} is also a common benchmark for DG, but not commonly used to evaluate sDG methods. The benchmark consists of 4 datasets (PASCAL-VOC, LabelMe, Caltech-101, SUN09) with differing styles with 5 classes. We train on the PASCAL-VOC domain and test the trained model on the remaining target domains.

\subsection{Data Augmentation}

In our experiments, we used the Random Augmentation \citep{cubuk2020randaugment} strategy as the augmentation function. The random augmentation method has two hyperparameters, the augmentation magnitude, and the number of transformations. Generally, previous works have used random augmentation by fixing the hyperparameters.  

As outlined in \Cref{alg:peer}, we periodically reinitialize the augmentation function by randomly selecting two hyperparameters, ensuring diverse augmented samples (\Cref{fig:otdd_small}). We find that changing the random augmentation configuration during training enhances generalization. Specifically, randomly selecting the parameters of the random augmentation (e.g., randomizing the augmentation magnitude in the torchvision implementation of \citep{cubuk2020randaugment}). While training a single model on these varied samples can lead to feature distortion, \textsc{peer} mitigates this through parameter averaging. In \Cref{sec:experiment}, we have shown that simple random augmentation outperforms sophisticated augmentation strategies devised for single source domain generalization.

\subsection{Baselines}

Here, we provide detailed descriptions of each baseline.
ERM \citep{ermkolt} is the baseline of training without data augmentation, followed by several augmentation-based sDG methods that use complex adversarial schemes to generate challenging augmentations \citep{qiao2020, wang2021, li2021}. 
M-ADA \citep{qiao2020} adopted a Wasserstein autoencoder to regularize perturbation in the latent space, L2D \citep{wang2021} takes a meta-learning approach to generate augmented domains, while PDEN \citep{li2021} and AdvST \citep{zheng2024advst} expand the training domains by progressively learning multiple augmentation modules, each simulating different domain shifts. Alternatively, MetaCNN \citep{wan2022} used a meta-convolutional network to learn generalized meta-features from local convolutional features. In contrast, we show that with \textsc{peer}, simple random augmentation can outperform all the baselines.

\begin{table*}[t!]
\caption{Target domain accuracy with different backbone architectures.}
\label{tab:accuracy_modelsize}
\centering
\begin{adjustbox}{width=0.8\textwidth}\setlength{\tabcolsep}{4pt}
\centering

\begin{tabular}{@{}lcccccccccccc@{}}
\toprule
& \multicolumn{4}{c}{{PACS}}  & \multicolumn{4}{c}{{Office-Home}} & \multicolumn{4}{c}{{VLCS}}\\
\cmidrule(lr){2-5} \cmidrule(lr){6-9} \cmidrule(lr){10-13}
{Method} & A & C & S & Avg. & Art & Clipart & Product & Avg. & L & C & S & Avg.\\

\midrule
& \multicolumn{4}{c}{{AlexNet}} & \multicolumn{4}{c}{{ResNet-18}} & \multicolumn{4}{c}{{ResNet-18}}\\
\cmidrule(lr){2-5} \cmidrule(lr){6-9} \cmidrule(lr){10-13}

RandAug [\citenum{cubuk2020randaugment}] & 54.17 & \textbf{47.48} & 65.11 & 55.59 & 43.10 & 45.47 & 61.67 & 50.01 & 57.58 & 93.18 & 66.56 & 72.44\\

PEER (ours) & \textbf{62.66} & 47.40 & \textbf{68.21} & \textbf{59.42} & \textbf{56.81} & \textbf{54.23} & \textbf{70.84} & \textbf{60.63} & \textbf{67.00} & \textbf{97.73} & \textbf{72.56} & \textbf{79.10}\\

\midrule
& \multicolumn{4}{c}{{ResNet-18}} & \multicolumn{4}{c}{{ResNet-50}} & \multicolumn{4}{c}{{ResNet-50}}\\
\cmidrule(lr){2-5} \cmidrule(lr){6-9} \cmidrule(lr){10-13}

RandAug [\citenum{cubuk2020randaugment}] &  65.64 & 38.27 & 56.32 & 53.68 & 64.11 & 53.86 & 76.70 &  64.89 & 56.95 & 94.39 & 71.09 & 74.15\\ 

PEER (ours) & \textbf{70.08} & \textbf{50.85} & \textbf{70.71} & \textbf{63.88} & \textbf{67.10} & \textbf{59.88} & \textbf{79.69} & \textbf{68.89} & \textbf{62.46} & \textbf{99.01} & \textbf{79.03} & \textbf{80.16}\\

\bottomrule
\end{tabular}

\end{adjustbox}
\end{table*}

\begin{table*}[ht!]
\caption{Variance of the target domain accuracy with backbone architectures.}
\label{tab:fluctuation_modelsize}
\centering
\begin{adjustbox}{width=0.8\textwidth}\setlength{\tabcolsep}{4pt}
\centering

\begin{tabular}{@{}lcccccccccccc@{}}
\toprule
& \multicolumn{4}{c}{{PACS}}  & \multicolumn{4}{c}{{Office-Home}} & \multicolumn{4}{c}{{VLCS}}\\
\cmidrule(lr){2-5} \cmidrule(lr){6-9} \cmidrule(lr){10-13}
{Method} & A & C & S & Avg. & Art & Clipart & Product & Avg. & L & C & S & Avg.\\

\midrule
& \multicolumn{4}{c}{{AlexNet}} & \multicolumn{4}{c}{{ResNet-18}} & \multicolumn{4}{c}{{ResNet-18}}\\
\cmidrule(lr){2-5} \cmidrule(lr){6-9} \cmidrule(lr){10-13}

RandAug [\citenum{cubuk2020randaugment}] & 2.23 & 4.81 & 5.01 & 4.02 & \textbf{3.49} & 2.17 & 2.74 & 1.89 & 3.02 & 1.61 & \textbf{1.96} & 2.20\\

PEER (ours)& \textbf{2.01}  & \textbf{3.98} & \textbf{4.77} & \textbf{3.59} & 3.99 & \textbf{1.41} & \textbf{1.80} & \textbf{1.31} & \textbf{2.05} & \textbf{1.61} & 2.10 & \textbf{1.92}\\

\midrule
& \multicolumn{4}{c}{{ResNet-18}} & \multicolumn{4}{c}{{ResNet-50}} & \multicolumn{4}{c}{{ResNet-50}}\\
\cmidrule(lr){2-5} \cmidrule(lr){6-9} \cmidrule(lr){10-13}

RandAug [\citenum{cubuk2020randaugment}] & 6.17 & 7.32 & \textbf{6.44} & 6.64 & 7.17 & \textbf{2.41} & 4.55 & 4.71 & 3.45 & 2.11 & \textbf{2.73} & 2.76 \\

PEER (ours) & \textbf{3.03} & \textbf{4.56} & 9.44 & \textbf{5.68} & \textbf{2.24} & 4.41 & \textbf{0.81} & \textbf{2.49} & \textbf{2.67}  & \textbf{1.72} & 3.57 & \textbf{2.65} \\

\bottomrule
\end{tabular}

\end{adjustbox}
\end{table*}

\begin{table*}[ht!]
\caption{Target domain accuracy with/without projection head $R$.}
\label{tab:accuracy_projection_head}
\centering
\begin{adjustbox}{width=0.77\textwidth}\setlength{\tabcolsep}{4pt}
\centering\begin{tabular}{@{}lcccccccccc@{}}
\toprule
& & \multicolumn{4}{c}{{PACS}} & \multicolumn{5}{c}{{Digits}}\\
\cmidrule(lr){3-6} \cmidrule(lr){7-11}
{Method} & Proj. Head. & A & C & S & Avg. & SVHN & M-M & S-D & USPS & Avg.\\
\midrule
PEER (ours) & \ding{51} & 62.66 & \textbf{47.40} & \textbf{68.21} & \textbf{59.42} & 70.79 & 76.84 &  \textbf{83.05} &  \textbf{93.57} &  \textbf{81.06}\\
PEER (ours) & \ding{55} & \textbf{62.76} & 43.26 & 66.00 & 57.34 & \textbf{76.34} & \textbf{93.07} & 68.96 & 80.36& 79.68\\
\bottomrule
\end{tabular}

\end{adjustbox}
\end{table*}

\subsection{Model Architecture}\label{appendix:architecture}

We report the details of model architectures used in our experiments. All models were built to match the architecture used in previous studies. 

\paragraph{Task Model}
The task model architecture varies in each experiment. For each experiment, we report the feature extractor $H$ and the regularization head $R$ of the task model $F$. Please note that the proxy model $P$ uses a model with an identical architecture as the task model $F$.

The task model used in the PACS experiment is AlexNet \citep{alexnet}, pre-trained on ImageNet \citep{imagenet}. The model consists of 5 convolutional layers with channels of \{96, 256, 384, 384, 256\}, followed by two fully-connected layers of size $4096$ units. The regularization head $R$ is a $3$ layer MLP. The output dimension of the regularization head is $1024$. 

The task model used in the Digits experiment is a multi-layer CNN network (i.e. {conv-pool-conv-pool-fc-fc-softmax}). The architecture consists of two 5 × 5 convolutional layers, with 64 and 128 channels respectively. Each convolutional layer is followed by a MaxPooling layer (2 × 2). The network also includes two fully connected layers with sizes of $1024$, $1024$ being the final output dimension of the feature extractor. The regularization head $R$ is a $2$ layer MLP. The output dimension of the regularization head is $128$.

Lastly, the task model used in the Office-Home and VLCS experiment is a ResNet-18 network. The ResNet is torchvision implemented and pre-trained on the ImageNet dataset. The regularization head $R$ is a $3$ layer MLP. The output dimension of the regularization head is $1024$.

\paragraph{Teacher Model for the PEER vs. Teacher Experiment}

For the PEER vs. Teacher experiment, we used pre-trained models as a teacher model. In the PACS experiment, we used a pre-trained RegNetY-16GF model. The RegNetY-16GF is a variant of the RegNet family, a line of foundation image models introduced in \citet{radosavovic2020designing} for image classification. The name of the model indicates its configurations, where the "Y" indicates the convolution method, and the "16GF" represents the model's capacity or complexity. We implement the model, and its model weights using the torchvision \citep{torchvision} library. For the Digits experiment, we used a pre-trained model sharing the same architecture as the task model. As elaborated in \Cref{appendix:model-to-model}, this is because a pre-trained model fit for use in digit classification was hard to obtain. Hence, following the practice of \citet{MIRO}, we trained the model with the source and target domains of Digits to create an Oracle model.

\subsection{Model Training}\label{appendix:training}
In this section, we elaborate on the details of the training process. We explicitly state the training hyperparameters (e.g., number of training epochs, augmentation reinitialization criteria $k$, learning rate, the type of the optimizer, learning rate scheduler, and batch size). 
All experiments are carried out using a single NVIDIA RTX 6000. 

\paragraph{PACS}
For the PACS experiment, we set the training epochs as $200$, and the augmentation reinitialization criteria $k$ as $10$. We tuned the number of epochs by analyzing the training behavior of the generators. We set the learning rate as $1e-4$, using the Adam optimizer \citep{kingma2017adam}.
The batch size was set as $128$. In total, the PACS experiment took roughly 101 minutes.

\paragraph{Digits}
For the Digits experiment, we set the training epochs as $1000$, and the augmentation reinitialization criteria $k$ as $10$. The learning rate was tuned as $0.0001$, using the Adam optimizer. The batch size was set as $128$. In total, the Digits experiment took roughly 233 minutes.

\paragraph{Office-Home}
For the Office-Home experiment, the training epochs are set as $200$, and the $k$ as $10$. The learning rate was set as $0.0001$, using the Adam optimizer. The batch size was set as $64$. In total, the Office-Home experiment took roughly 128 minutes.

\paragraph{VLCS}
Lastly, for the VLCS experiment, we train for $200$ epochs, and the $k$ as $10$. The learning rate was set as $0.0001$, using the Adam optimizer. The batch size was set as $128$. In total, the VLCS experiment took roughly 117 minutes.

\subsection{Model pre-training}\label{appendix:pretraining}
In this section, we report the information regarding the pre-training process. As mentioned above, we pre-trained our task model with the source domain before the main training procedure. We announce the number of pre-training epochs, the learning rate, the optimizer, the learning rate scheduler, and the batch size.

\paragraph{PACS}
We pre-trained the AlexNet with the train data of the Photo domain, using the train split introduced in the original paper \citep{pacs}. We pre-trained the model for $60$ epochs, with a learning rate of $0.005$ using the SGD optimizer. We further used the Step learning rate scheduler with a gamma rate (i.e. the strength of the learning rate decay) of $0.5$. The batch size was set as $32$. 

\paragraph{Digits}
For the Digits experiment, we set the number of pre-training epochs as $100$, with a learning rate of $0.0001$ using the Adam optimizer. The batch size was set as $256$.

\paragraph{Office-Home}
We pre-trained the ResNet18 with the train split of the Real World domain. We pre-trained the model for $100$ epochs, with a learning rate of $0.0001$ using the Adam optimizer. We used no learning rate scheduler. The batch size was set as $64$. 

\paragraph{VLCS}
We pre-trained the ResNet18 with the train split of the PASCAL VOC domain. We pre-trained the model for $100$ epochs, with a learning rate of $0.0001$ using the Adam optimizer. We used no learning rate scheduler. The batch size was set as $64$. 
 
\subsection{Hyperparameters}\label{appendix:hyperparameters}

In this part, we state the hyperparameters used in our experiments. 

$\lambda$ is a balancing coefficient for $L_{\textsc{peer}}$, an objective adopting the feature-decorrelation loss introduced in \citet{barlow_twins}. We tuned $\lambda$ using experimental results of the original paper and \citet{tsai2021}. In the original paper, the author reported the optimal value of the balancing term as $0.005$, which remains consistent under varying projection dimensions. We set this as a starting point for hyperparameter tuning. We find that if $\lambda$ balances the off-diagonal term (i.e. redundancy reduction term) and the diagonal term (i.e. alignment term) to a similar degree, no significant differences are observed. Furthermore, switching $\lambda$ to $\frac{1}{d} \approx 0.0001$ showed no significant changes to the learning process. Here, $d$ denotes the projection dimension of the regularization head $\mathcal{R}$ (regularization head output space). While we cannot guarantee an optimal value for $\lambda$, we set $\lambda=0.005$ for our experiments using \textsc{peer}. 

$k$ is an augmentation reinitialization criterion that performs two roles. (1) Augmentation reinitialization: For every $k$ epoch, the augmentation function is initialized. Here, reinitialization refers to the change in augmentation policy. For instance, for random augmentation, reinitialization refers to the change in augmentation strength. Alternatively, for augmentation techniques that utilize a learnable module \citep{li2021}, the reinitialization would refer to reinitializing the parameters of the augmentation module. The motive behind the reinitialization is to expose the proxy model with diverse augmentations, (2) \textsc{peer} update: For every $k$ epoch, the parameters of the proxy model $P$ are used to update the task model by averaging their parameters.

Lastly, $w$ is a hyperparameter used in \Cref{loss:f}, which balances the ERM objective and the regularization objective \Cref{loss:peer}. As studied in \Cref{appendix:hyperparameter_experiment}, $w$ does not affect the performance of our method. We have set $w$ as $2.0$ based upon experimental results in \Cref{tab:ablation_PACS_combined}.

\section{Licenses for Existing Assets}

    In the process of performing our research, many existing assets were used. For the implementation of the models and their weights, we have used the torchvision library \citep{torchvision} (BSD License). We also made sure that the datasets used in our experiments were open-source public datasets that do not pose license issues. Specifically, we use data collected from multiple sources: torchvision, Dassl (https://github.com/KaiyangZhou/Dassl.pytorch), huggingface (https://huggingface.co/datasets), and from the original papers. We made sure to cite the authors for their contribution to datasets and benchmarks. We list the license types of each dataset, in cases where we could retrieve them. For instance, PACS \citep{pacs} uses the CC BY 4.0 license. Digits \citep{deng2012mnist} uses the Creative Commons Attribution-Share Alike 3.0 license. Office-Home \citep{officehome} uses a custom license that allows non-commercial research and educational purposes. VLCS \citep{fang2013unbiased} uses a custom license (http://host.robots.ox.ac.uk/pascal/VOC/).

\end{document}